\documentclass[sw]{iosart2x_arxiv}

\usepackage{graphicx}
\usepackage[utf8]{inputenc}
\usepackage{stmaryrd} 
\usepackage{listings}
\usepackage{menukeys}
\usepackage{booktabs}
\usepackage{bm} 
\usepackage{enumitem}
\usepackage{balance}

\usepackage{tikz}
\usepackage{collcell}
\usepackage{multirow}

\newcommand{\tabkg}{\textit{Tab2KG}}
\newcommand{\dslless}{\textit{DSL*}}
\newcommand{\dsl}{\textit{DSL}}
\newcommand{\tkmatch}{{\textit{T2KMatch}}}
\newcommand{\voc}[2]{\texttt{#1:\allowbreak #2}}
\newcommand{\schema}[1]{\texttt{#1}}

\newtheorem{definition}{Definition} 

\lstdefinelanguage{ttl} {
    alsoletter={-},
    morekeywords={a,owl,ex,rdf,rml,sosa,csvw,rr}
}

\newcommand*{\MinNumber}{0.88}
\newcommand*{\MaxNumber}{0.952}%

\newcommand{\ApplyGradient}[1]{%
    \pgfmathsetmacro{\PercentColor}{max(min(100.0*(#1 - \MinNumber)/(\MaxNumber-\MinNumber),100.0),0.00)} %
    \hspace{-0.33em}\colorbox{lightgray!\PercentColor!white}{#1}
}

\newcolumntype{R}{>{\collectcell\ApplyGradient}c<{\endcollectcell}}

\pubyear{0000}
\volume{0}
\firstpage{1}
\lastpage{1}

\begin{document}

\makeatletter
\let\put@numberlines@box\relax
\makeatother

\begin{frontmatter}

\runtitle{\tabkg{}}

\title{\tabkg{}: Semantic Table Interpretation with Lightweight Semantic Profiles}

\author[A]{\inits{S.}\fnms{Simon} \snm{Gottschalk}\ead[label=e1]{gottschalk@L3S.de}%
\thanks{Corresponding author. \printead{e1}.}}
and
\author[B]{\inits{E.}\fnms{Elena} \snm{Demidova}\ead[label=e2]{elena.demidova@cs.uni-bonn.de}}

\address[A]{L3S Research Center, \orgname{Leibniz Universität Hannover},
Hannover, \cny{Germany}\printead[presep={\\}]{e1}}
\address[B]{\orgname{Data Science \& Intelligent Systems (DSIS), University of Bonn},
Bonn, \cny{Germany}\printead[presep={\\}]{e2}}


\begin{abstract}
Tabular data plays an essential role in many data analytics and machine learning tasks. Typically, tabular data does not possess any machine-readable semantics. In this context, semantic table interpretation is crucial for making data analytics workflows more robust and explainable. This article proposes Tab2KG -- a novel method that targets at the interpretation of tables with previously unseen data and automatically infers their semantics to transform them into semantic data graphs. 
We introduce original lightweight semantic profiles that enrich a domain ontology's concepts and relations and represent domain and table characteristics. 
We propose a one-shot learning approach that relies on these profiles to map a tabular dataset containing previously unseen instances to a domain ontology. 
In contrast to the existing semantic table interpretation approaches, Tab2KG relies on the semantic profiles only and does not require any instance lookup. This property makes Tab2KG particularly suitable in the data analytics context, in which data tables typically contain new instances.
Our experimental evaluation on several real-world datasets from different application domains demonstrates that Tab2KG outperforms state-of-the-art semantic table interpretation baselines.
\end{abstract}


\end{frontmatter}


\section{Introduction}
\label{sec:introduction}

A vast amount of data is currently published in a tabular format \cite{cafarella2018ten,ritze2017matching,mitlohner2016characteristics}. Typically, this data does not possess any machine-readable semantics. Semantic table interpretation is an essential step to make this data usable for a wide variety of applications, with data analytics workflows (DAWs) as a prominent example \cite{gottschalk2019simple}. DAWs include data mining algorithms and sophisticated deep learning architectures, and require a large amount of heterogeneous data as an input. 
Typically, DAWs treat tabular data as character sequences and numbers 
and potentially miss important information that has not been made available explicitly. This practice can lead to error-prone analytics processes and results, particularly when data analytics frameworks utilize the data from various heterogeneous sources. Therefore, DAWs can substantially profit from the semantic interpretation of the involved data tables \cite{garda2019semantics, pomp2018applying}. 

In this context, semantic table interpretation aims to transform the input data table into a semantic data graph. In this process, table columns are mapped to a domain ontology's classes and properties; table cell values are transformed into literals, forming the data graph -- a network of semantic statements, typically encoded in RDF. 
In the context of DAWs, semantic table interpretation can bring several advantages. First, an abstraction from tabular data to semantic concepts and relations can guide domain experts in the DAW creation \cite{gottschalk2019simple}. Second, validation options (e.g., type inference) that become available through the semantic table interpretation can increase the robustness of DAWs \cite{hartenfels2017type}. Third, semantic descriptions can be employed to facilitate the explainability of the data analytics results \cite{lecue2019role}. Finally, semantic table interpretation adds structure directly usable for knowledge inference \cite{lehmann2017distributed}.

The existing approaches to semantic table interpretation do not adequately support the interpretation of tabular data for DAWs. 
At the core of such approaches (e.g., \cite{MTab,cremaschi2020fully,chen2019colnet}) is the instance lookup task, where table cells are linked to known instances in a target knowledge graph, with DBpedia \cite{auer2007dbpedia} being a popular cross-domain target. Subsequent steps such as property mapping are based on the results of this lookup step. However, as shown by Ritze et al. \cite{ritze2016profiling}, only about $3\%$ of the tables contained in the 3.5 billion HTML pages of the Common Crawl Web Corpus\footnote{\url{http://commoncrawl.org/}} can be matched to DBpedia. In the context of DAW, the input data typically represents new instances (e.g., sensor observations, current road traffic events, \ldots), and substantial overlap between the tabular data values and entities within existing knowledge graphs cannot be expected.

In this article, we present \tabkg{} -- a novel semantic table interpretation approach. \tabkg{} aims to transform a data table into a semantic data graph. 
As a backbone of the data graph, \tabkg{} relies on an existing domain ontology that defines the concepts and relations in the target domain.
To facilitate the transformation, \tabkg{} introduces original lightweight semantic profiles for domains and data tables.
Domain profiles enrich ontology relations and represent domain characteristics. A domain profile associates relations with feature vectors representing data types and statistical characteristics such as value distributions.
To transform a data table, \tabkg{} first creates a data table profile. 
Then, \tabkg{} uses the domain and data table profiles to transform the table into a data graph using a novel one-shot learning approach.

Lightweight semantic profiles generated by \tabkg{} can be utilized as compact domain representations
and complement and enrich existing dataset catalogs. Such profiles can be generated automatically from the existing datasets and described using the DCAT\footnote{\url{https://www.w3.org/TR/vocab-dcat-2/}} and the SEAS\footnote{\url{https://ci.mines-stetienne.fr/seas/index.html}} vocabularies to enable their reusability. We believe that lightweight semantic profiles presented in this article are an essential contribution that can benefit a wide range of semantic applications beyond semantic table interpretation.

In summary, \tabkg{} is driven by the following three goals, which distinguish \tabkg{} from traditional semantic table interpretation approaches:
\begin{itemize}
    \item Interpretation of previously unseen data: \tabkg{} can interpret domain-specific tables in the common case where the table values are not yet present in existing knowledge graphs.
    \item Usage of lightweight semantic profiles: \tabkg{} relies on lightweight semantic profiles with profile features which can be easily modeled as part of data catalogs (e.g., histograms and data types).
    \item Generalisation through one-shot learning: \tabkg{} generalises towards new domain ontologies with a one-shot learning approach.
\end{itemize}

Consequently, our contributions presented in this article are as follows:

\begin{enumerate}
    \item We introduce lightweight semantic domain and table profiles. Domain profiles enrich relations of domain ontologies and serve as a lightweight domain representation. Semantic table profiles summarize data tables facilitating effective semantic table interpretation.
    \item We propose the \tabkg{} approach to transform tabular data into a data graph with one-shot learning based on semantic profiles. 
    \item We evaluate the proposed method on several real-world datasets.
    Our evaluation results demonstrate that \tabkg{} outperforms state-of-the-art semantic table interpretation baselines. 
    \item We make the scripts for creating lightweight semantic profiles and transforming data tables into data graphs publicly available\footnote{\url{https://github.com/sgottsch/Tab2KG}}.
\end{enumerate}

The structure of this article is as follows: In Section \ref{sec:example}, we introduce a running example used throughout this article. Then, in Section \ref{sec:problem_statement}, we define the problem of semantic table interpretation, followed by the definition and creation of domain and data table profiles (Section \ref{sec:profiles}).
In Section \ref{sec:approach}, we describe our proposed \tabkg{} approach and its implementation (Section \ref{sec:implementation}).
We present evaluation setup and results in Sections \ref{sec:evaluation} and \ref{sec:evaluation_results}, followed by a discussion of our profile-based approach in Section \ref{sec:discussion}. Then, we discuss related work in Section \ref{sec:related_work}. Finally, we provide a conclusion in Section \ref{sec:conclusion}.

\section{Running Example from the Weather Observation Domain}
\label{sec:example}

As a running example, we use the weather observation domain and a data table that provides observations of sensors measuring air conditions.

\begin{figure*}[ht]
    \centering
    \begin{minipage}[t]{0.4\textwidth}
        \centering
        \includegraphics[width=\textwidth]{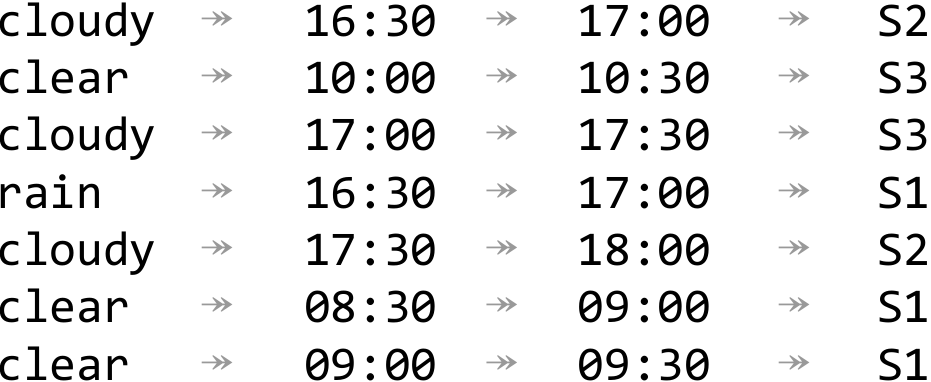} 
        \caption{Example of a data table as a tab-separated file without column titles.}
        \label{fig:example_table}
    \end{minipage}\hfill
    \begin{minipage}[t]{0.46\textwidth}
        \centering
        \includegraphics[width=\textwidth]{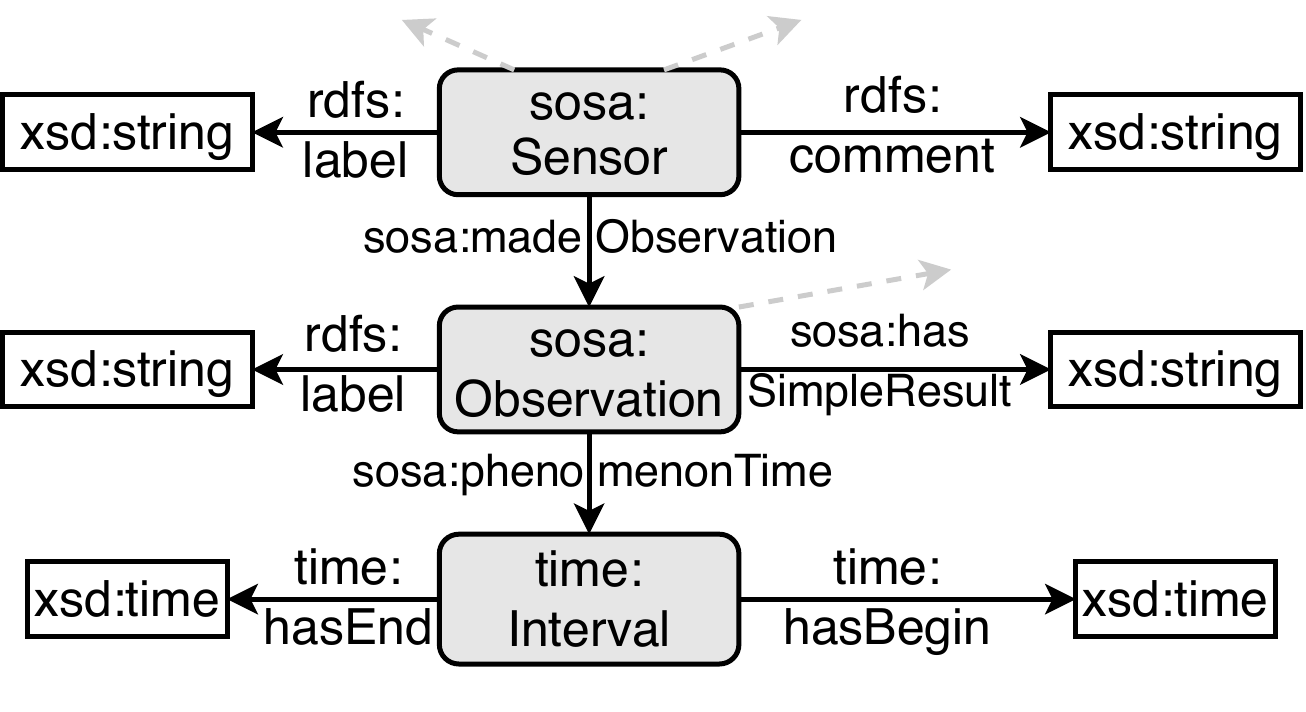} 
        \caption{Excerpt of the Semantic Sensor Network Ontology.}
        \label{fig:example_schema}
    \end{minipage}
\end{figure*}

Consider the table in Fig. \ref{fig:example_table} that contains weather observation sensor data, separated by a tab character ($\twoheadrightarrow$). The table does not include column titles. As a human, we can observe that the first column refers to the air condition (\textit{cloudy}, \textit{clear}, \textit{rain}). The second and third column may represent a time interval of the measurement (e.g., \textit{16:30} and \textit{17:00}). The fourth column 
containing the values \textit{S2}, \textit{S3}, and \textit{S1} is hard to interpret without background knowledge.

A domain ontology and a domain profile are required to map the table columns to their respective semantic concepts and to transform the whole table into a data graph.

\begin{figure}[hb]
   \centering
       \includegraphics[width=\linewidth]{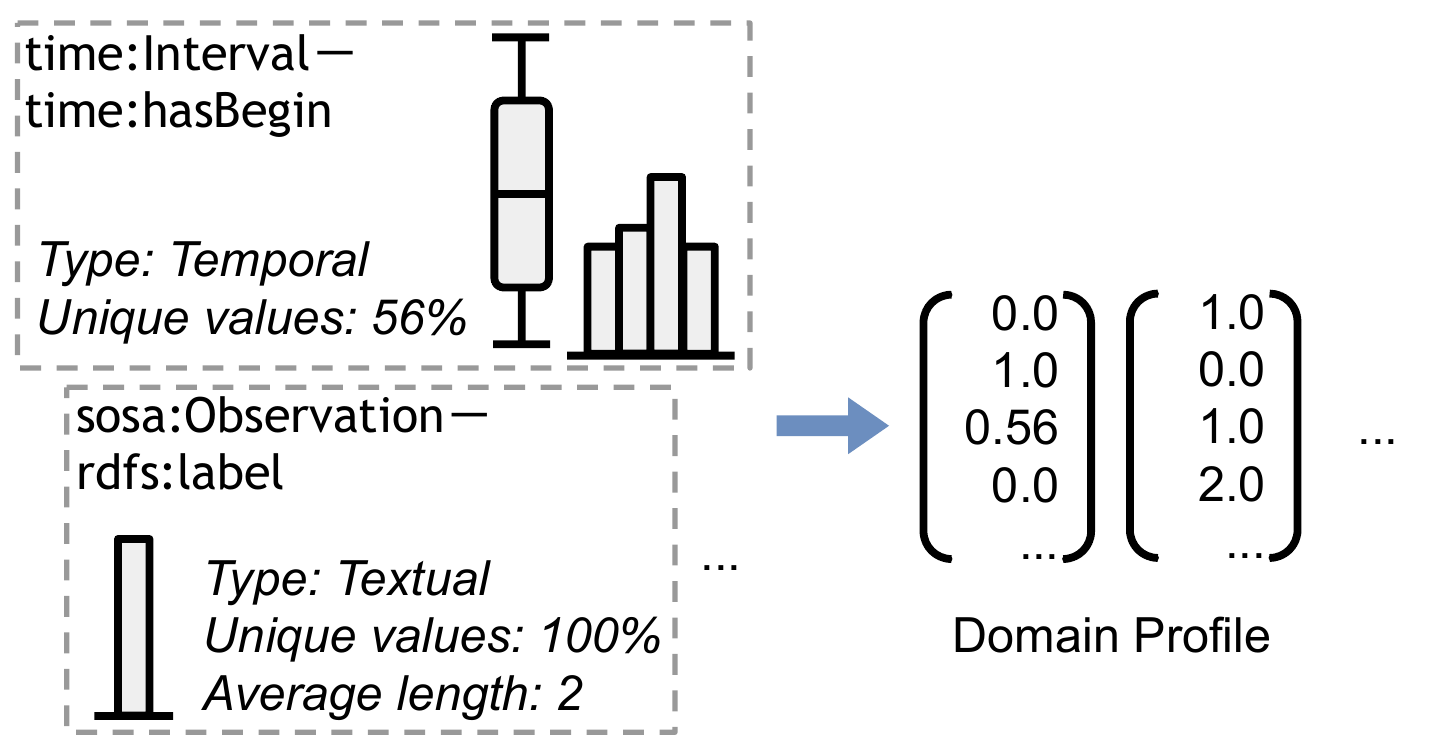} 
 \caption{An example profile of the weather observation domain. 
The domain profile is represented as a set of feature vectors, each containing statistical features, such as value distributions. 
The domain profile can also be used for visualization.}
 \label{fig:domain_profile_example}
\end{figure}

\begin{figure*}[ht]
    \centering
    \begin{minipage}[t]{0.44\textwidth}
        \centering
        \includegraphics[width=\textwidth]{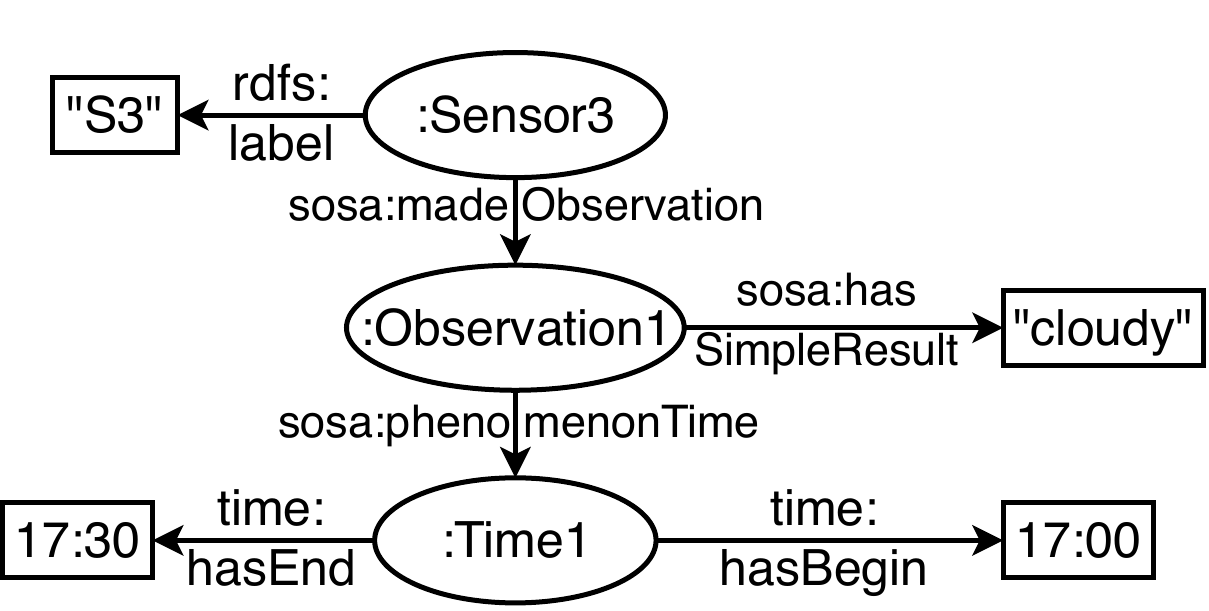} 
        \caption{Correct mapping of the third line in Fig.~\ref{fig:example_table} to the ontology in Fig.~\ref{fig:example_schema}. For brevity, we omit \voc{rdf}{type} relations.}
        \label{fig:correct_mapping_example}
    \end{minipage}\hfill
    \begin{minipage}[t]{0.44\textwidth}
        \centering
        \includegraphics[width=\textwidth]{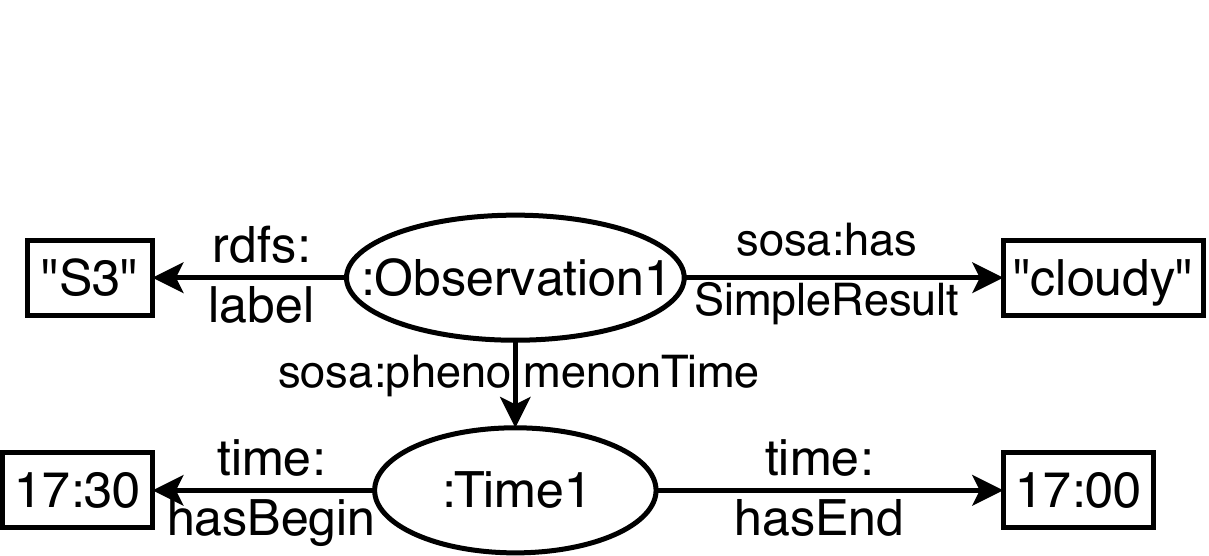} 
        \caption{Incorrect mapping of the third line in Fig.~\ref{fig:example_table} to the ontology in Fig.~\ref{fig:example_schema}. For brevity, we omit \voc{rdf}{type} relations.}
        \label{fig:wrong_mapping_example}
    \end{minipage}
\end{figure*}

\begin{itemize}
    \item We use the Semantic Sensor Network Ontology\footnote{\texttt{sosa:} \url{https://www.w3.org/TR/vocab-ssn/}} illustrated in Fig. \ref{fig:example_schema} as \textbf{domain ontology}. Among others, this ontology provides classes to model sensors, their observations, and corresponding time intervals.
    \item Fig. \ref{fig:domain_profile_example} provides an exemplary illustration of a \textbf{domain profile} in the weather observation domain. Here, we illustrate statistical features of two observation properties (the beginning of the observation and the sensor label) using box plots and histograms.
\end{itemize}

\tabkg{} transforms the table into the data graph shown in Fig. \ref{fig:correct_mapping_example}. As we can observe, the first three columns are mapped to the observations and their time intervals. The fourth column is mapped to the sensor labels. Note that not all properties of the example domain ontology are covered by the data graph, as the domain ontology contains relations not present in the table (e.g., \voc{sosa}{Observation} - \voc{rdfs}{label} - \voc{xsd}{string}).

The transformation process is challenging and potentially error-prone. For example, Fig. \ref{fig:wrong_mapping_example} illustrates a wrong transformation result, with an incorrect column mapping and an erroneous graph structure. In this case, the sensor label ``S3'' was erroneously interpreted as an observation label. In addition, the beginning and end times are swapped. \tabkg{} utilizes semantic profiles to avoid such interpretation errors.

\section{Problem Statement}
\label{sec:problem_statement}

In this section, we first formally define relevant concepts. Then, we present the task of semantic table interpretation.

The entities and relations in the domain of interest (e.g., weather observation or soccer) can be represented in a domain knowledge graph.

\begin{definition}
  A \textbf{domain knowledge graph} is a graph $G = (N,R)$, whose nodes $N$ represent entities and literals, and whose edges $R$ represent relations between these nodes in the specific domain. 
\end{definition}

A domain knowledge graph consists of two sub graphs: a domain ontology and a data graph. 

\begin{definition}
A \textbf{domain ontology} $G_{O}=\allowbreak(N_{O},\allowbreak R_O)$,
$N_{O} \subset N, R_{O} \subset R$, where $N_{O}=C \cup D \cup P$ includes a set of classes $C$, a set of data types $D$, and a set of properties $P = P_d \cup P_o$ relevant in the specific domain, where $P_d$ are data type properties, and $P_o$ are object properties. Data type properties relate entities to literals. Object properties relate entities to each other.

The relations represented by $R_{O} = R_{OC} \cup R_{OD} $ include class relations $R_{OC}$ and data type relations $R_{OD}$:

\begin{itemize}
\item $R_{OC}$ is the set of class relations:

$R_{OC} \subseteq C \times P_o \times C$.
\item $R_{OD}$ is the set of data type relations:

$R_{OD} \subseteq C \times P_d \times D$.
\end{itemize}
\end{definition}

For example, in the excerpt of Semantic Sensor Network Ontology illustrated in Fig. \ref{fig:example_schema}, 
(\voc{sosa}{Sensor} \voc{sosa}{madeObservation} \voc{sosa}{Observation}) is a class relation and
(\voc{sosa}{Sensor} \voc{rdfs}{label} \voc{xsd}{string}) is a data type relation.

\begin{definition}
\label{def:datagraph}
A \textbf{data graph} is a graph $G_D=(N_{D},R_{D})$, $N_{D} \subset N, R_{D} \subset R$. 
The nodes $N_{D} = C \cup  D \cup  E \cup L$ include classes $C$, data types $D$, entities $E$ and literals $L$. Each literal $l \in L$ is assigned a data type $dt(l) \in D$. Within $R_D$, we distinguish between entity relations ($E \times P_o \times E$) and literal relations ($E \times P_d \times L$).

\end{definition}

A data table is defined as follows:

\begin{definition}
\label{def:table}
  A \textbf{data table} $T$ is a $M \times N$ matrix consisting of $M$ rows and $N$ columns. 
  A cell $T_{m,n}$, $m \in \{1,\dots,M\}, n \in \{1,\dots,N\}$, represents a data value. 
  A row $r_m$ is a tuple that represents a set of semantically related entities.
A column $c_n$ represent a specific characteristic of the entities in a row.
\end{definition}

For example, the data table illustrated in Fig. \ref{fig:example_table} contains $M=7$ rows and $N=4$ columns, where the columns represent the weather conditions, time intervals, and sensor labels, and each row contains three semantically related entities: an observation, a time interval, and a sensor.
The column values can belong to different data types, including text, numeric, Boolean, temporal and spatial.

Semantic table interpretation is the task of transforming a data table into a data graph.  

\begin{definition}\label{def:semantification}
\textbf{Semantic table interpretation}: Given a data table $T$ and a domain knowledge graph $G$, create a data graph $G_D^T = (N_D^T, R_D^T)$ with nodes $N_D^T$ and relations $R_D^T$. Its literal values $L^T \subseteq N_D^T$ are connected to the entities $E^T \subset N^T_D$ and represent 
the values in the literal columns of $T$. 
The entities $E^T$ are connected via entity relations in $R_{D}^T$. 
\end{definition}

\section{Semantic Profiles}
\label{sec:profiles}

Semantic table interpretation in \tabkg{} does not require any instance lookup in a domain knowledge graph.
Instead, \tabkg{} uses a domain knowledge graph to create a lightweight semantic domain profile.
This domain profile, together with a domain ontology, builds reusable \textbf{domain background knowledge} that is 
later on used to interpret the data tables semantically.

The domain profile represents the domain knowledge graph regarding the value distributions. 
Note that the entities and literals in the domain knowledge graph do not need to overlap with the data tables' instances to be interpreted.

\tabkg{} involves the creation of two types of profiles: \textit{domain profiles} and \textit{data table profiles}, both represented as feature vectors and described in a semantic data catalog to facilitate their reusability. Such profiles are inspired by the dataset profiles described in \cite{ben2018rdf}, where statistical features are defined as an important element of a dataset profiles taxonomy. In \tabkg{}, the primary purpose of the domain and data table profiles is to enable effective and efficient access to the domain and table statistics for semantic table interpretation.

We present domain profiles in Section \ref{sec:semantic-profiles} and data table profiles in Section 
\ref{sec:semantic-profiles-data-table}.
We discuss profile features in Section \ref{sec:profile-features}.
Then, in Section \ref{sec:semantic-domain-and-data-table-profiles}, we describe how we represent domain and data table profiles in a semantic, machine-readable way.
Finally, in Section \ref{sec:catalog-example} we provide an example of a data catalog that includes semantic profiles.

\subsection{Domain Profiles}
\label{sec:semantic-profiles}

For creating a domain profile, we make use of a domain knowledge graph $G$ that contains representative values for the data type relations in the target domain. 
A \textbf{domain profile} $\Pi(G_O) = \{(r_D,\pi(r_D)) | r_D \in  R_{OD}\}$ of a domain ontology $G_O = (N_O, R_{OC} \cup R_{OD})$ is a set of data type relation profiles $\pi(r_D)$ derived from $G$, where a data type relation profile is a set of features of the literals covered by this data type relation in $G$'s data graph $G_D$.

\begin{definition}
The \textbf{data type relation profile} $\pi(r_D) \in \mathbb{R}^f$ of the data type relation $r_D \in R_{OD}$ is a vector that includes $f$ statistical characteristics (features) of the literal relations covered in the domain knowledge graph $G$.
\end{definition}

In brief, the profile of a data type relation $r_D$ is a feature vector containing a set of statistics, computed using all literals corresponding to $r_D$. A description of the features of the literal relations follows in Section~\ref{sec:profile-features}.

To create a profile for the data type relation $r_D = (c, p_d, d)$, we utilize all literals  $l\in G_{D}$, such that: $(e, p_d, l) \in R_{D}$, $(e, \texttt{rdf:type}, c) \in R_{D}$, and $dt(l)=d$. 

In our running example, in Fig. \ref{fig:correct_mapping_example}, the data type relation (\voc{sosa}{Sensor} \voc{rdfs}{label} \voc{xsd}{string}) in the domain ontology corresponds to the literal relation (\voc{}{Sensor3} \voc{rdfs}{label} ``S3'') in the data graph.  
Therefore, we use ``S3'' as one of the literals to create the data type relation profile.

\subsection{Data Table Profiles}
\label{sec:semantic-profiles-data-table}

To facilitate semantic interpretation of a data table, we create a data table profile. 

A \textbf{data table profile} is a set of column profiles, each representing a specific data table column.
More formally, the profile of a data table $T$ consists of a column profile $\pi(c_n), n \in \{1,\dots,N\}$ for each table column $c_n \in T$ as defined in Definition \ref{def:table}. 

A column profile is defined as follows:

\begin{definition}
A \textbf{column profile} $\pi(c_n) \in \mathbb{R}^f$ of a column $c_n$ is a vector of $f$ features of the values contained in that column. 
\end{definition}

We create column profiles using literal values contained in the table columns.  

Column profiles and data type relation profiles are created analogously and contain the same features, presented in Section \ref{sec:profile-features}.

\subsection{Profile Features}
\label{sec:profile-features}

Motivated by the RDF profile characteristics defined by Ellefi et al.~\cite{ben2018rdf},
we include data types, as well as completeness and statistical features
described in the following into the profiles in \tabkg{}.
The selection is motivated by the expected feature effectiveness for semantic table interpretation, i.e., matching the domain and data table profiles. 
We demonstrate in our evaluation that these features can facilitate an effective matching in several application domains.
This feature set can be extended to include relevant domains-specific characteristics.

\begin{itemize}
    \item \textbf{Data type:} We represent data types as binary profile features. We include fine-granular data types to facilitate the precise matching of domain and data table profiles. The following data type taxonomy includes the most common cases observed in our evaluation domains.
    \begin{itemize}[label=$\bullet$]
        \item \textbf{Text:} Categorical, URL, Email, Other
        \item \textbf{Numeric:} Integer, Decimal /  Sequential, Categorical, Other\footnote{following the taxonomy defined in \cite{alobaid2019typology}}
        \item \textbf{Boolean}
        \item \textbf{Temporal:} Date, Time, Date Time
        \item \textbf{Spatial:} Point, Linestring, Polygon
    \end{itemize}
    A data type relation or column can be assigned multiple (fine-granular) data types (e.g., integer and categorical). We provide technical details regarding the identification of fine-granular data types later in Section \ref{sec:implementation}.
    \item \textbf{Completeness:} We include the number of values, the number of non-null values and the number of distinct values as a completeness indicator.
    \item \textbf{Basic statistics:} For numeric values, we include the standard deviation, mean, skewness, kurtosis and number of outliers computed using the interquartile range (1.5 IQR) rule, as well as the average numbers of characters, digits, tokens, capital letters and special characters for the literals.
    \item \textbf{Histograms:} Histograms are an effective means for RDF data summarization \cite{harth2010data}. We create a histogram for a given number of buckets as part of the data type relation profile or column profile. As features, we add the number of literals in each bucket, in the increasing order of bucket ranges. For histogram creation, we remove the outliers detected before.
    \item \textbf{Quantiles:} We add quartiles and deciles to the profile (including minima and maxima).
\end{itemize}

To derive numerical features, we transform literals into numbers. The features of textual data type relations are computed based on the textual value lengths. Temporal values are transformed into timestamps. For spatial values, we consider the line string length or the polygon area, respectively.

\subsection{Profile-based Semantic Table Interpretation}
\label{sec:profile-based-sti}

Following the definition of domain profiles, their features and representation, we can now refine Definition~\ref{def:semantification}. Instead of requiring a data graph $G$, \textit{profile-based semantic table interpretation} solely requires a domain profile $\Pi(G_O)$.

\begin{definition}\label{def:profile_sti}
\textbf{Profile-based semantic table interpretation} is the task of semantic table interpretation (Definition~\ref{def:semantification}) of a data table $T$, given a domain ontology $G_O = (N_O, R_{OC} \cup R_{OD})$ and a domain profile $\Pi(G_O) = \{(r_D,\pi(r_D)) | r_D \in  R_{OD}\}$.
\end{definition}

The domain profile is typically built from a domain knowledge graph $G$. As soon as the domain profile is created, the data graph of $G$ is not required anymore for profile-based semantic table interpretation.

\subsection{Semantic Profile Representation}
\label{sec:semantic-domain-and-data-table-profiles}

Domain and data table profiles can be represented as \textbf{semantic profiles} in RDF, as an extension of the Data Catalog Vocabulary (DCAT)\footnote{\url{https://www.w3.org/TR/vocab-dcat-2/}} and the SEAS Statistics ontology\footnote{\url{https://ci.mines-stetienne.fr/seas/StatisticsOntology}}.

Within the DCAT vocabulary, a data catalog (\voc{dcat}{Data\-Catalog}) consists of datasets (\voc{dcat}{Da\-ta\-set}), 
where a dataset is a collection of data, published or curated by a single agent.
In the context of \tabkg{}, both the domain knowledge graph and the data tables can be represented using \voc{dcat}{Dataset}.
%
We extend the descriptions of datasets in a \tabkg{} data catalog to include semantic profiles. For example, we introduce an \schema{Attribute} class representing the data table columns and data type relations. We make the definitions of this vocabulary available online\footnote{\url{https://github.com/sgottsch/Tab2KG}}.

\begin{figure}
  \centering
    \includegraphics[width=\linewidth]{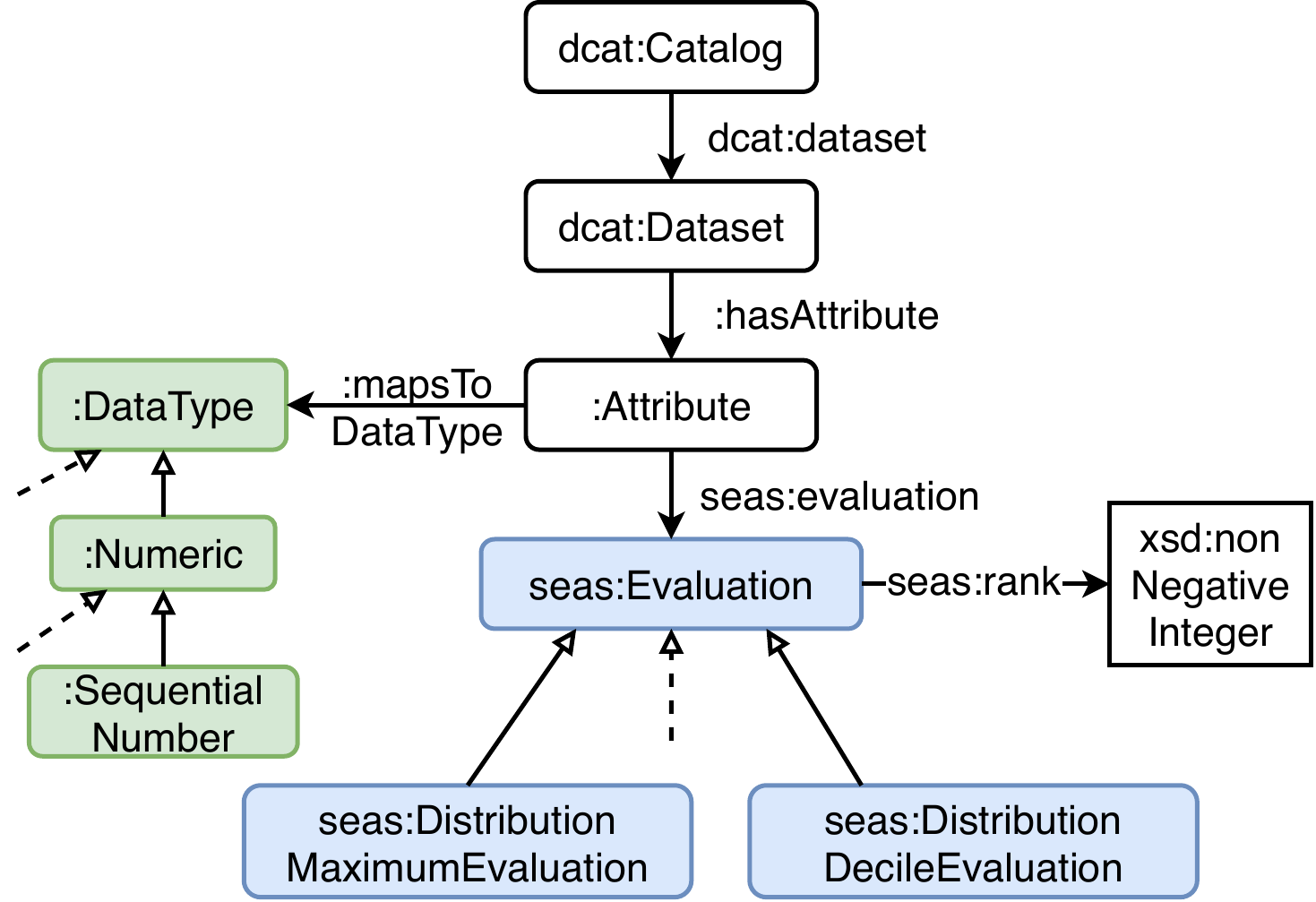}
      \caption{Classes and properties used for describing a semantic profile. $\rightarrowtriangle$ marks \voc{owl}{subClassOf} relations. Dashed arrows indicate the existence of further classes which are not included in this excerpt. The two feature types (data types and numeric features) are grouped by different colors.}
  \label{fig:semantic_profile}
\end{figure}

Fig. \ref{fig:semantic_profile} provides an overview of the classes involved in representing semantic profiles. 
A \voc{dcat}{Data\-set} in a \tabkg{} data catalog includes several attributes. 
In the case of a data table profile, these attributes are the columns. In the case of a domain profile, these attributes are the data type relations. 
These attributes are assigned the profile features, as presented in Section \ref{sec:profile-features}:
\begin{enumerate}
    \item Data types: Data type assignments follow the previously mentioned taxonomy.
\item Numeric features: The numeric profile features are represented through subclasses of \voc{seas}{Evaluation}. In the case of quartiles, deciles, and histograms, the values come with a rank. For example, we can denote the second decile using \texttt{seas:rank 2}.
\end{enumerate}

In the case of domain profiles, the existing mapping to the domain ontology can be modeled by connecting attributes to their corresponding classes, and data type properties \cite{gottschalk2019simple}.

Note that such semantic profiles do not only enable profile-based semantic table interpretation but can also be used to provide lightweight dataset visualizations, e.g., through box plots (quartiles) or as histograms. 

\subsection{Running Example: Weather Data Catalog}
\label{sec:catalog-example}

\begin{figure*}
  \centering
    \includegraphics[width=0.8\linewidth]{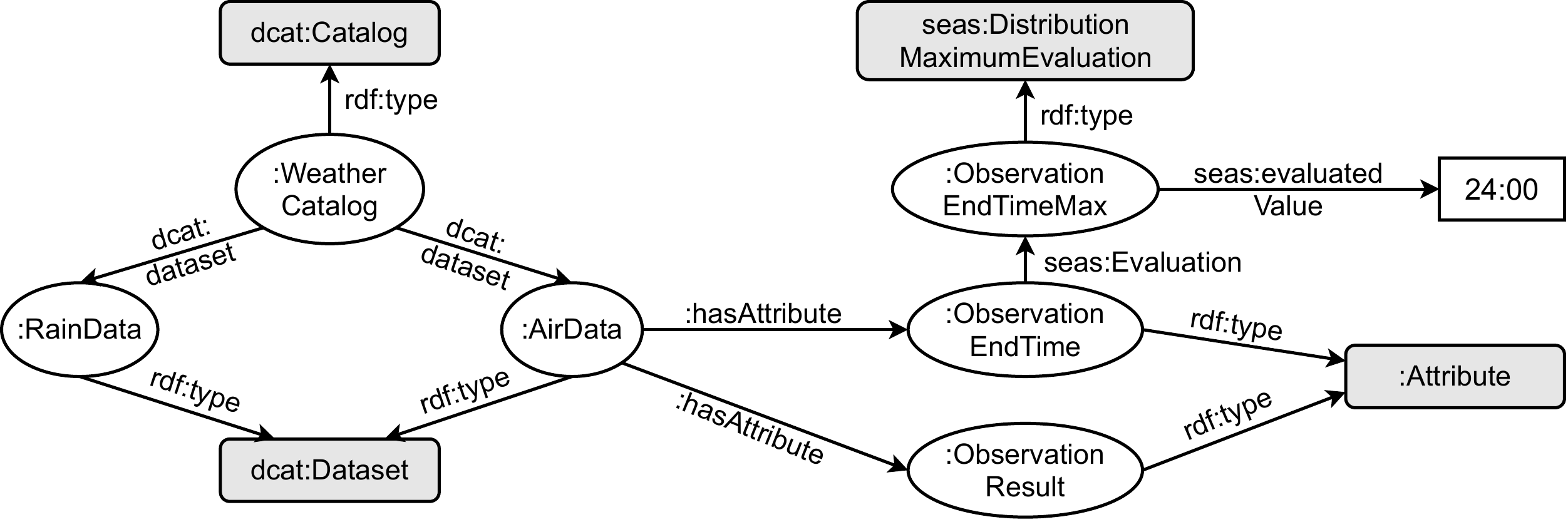}
      \caption{Running example: Excerpt of a weather data catalog containing two data tables and an exemplified column profile feature denoting the maximum end time value (\schema{ObservationEndTimeMax}).}
  \label{fig:data_catalog}
\end{figure*}

An excerpt of an example data catalog for our running example from the weather observation domain introduced in Section \ref{sec:example} is shown in Fig. \ref{fig:data_catalog}. The data catalog identified as \schema{WeatherCatalog} includes two data tables (\schema{RainData} and \schema{AirData}). Here, the \schema{AirData} data table has two columns, one of them with a column profile feature denoting the maximum value of the observation end time.

With \tabkg{}, we can directly utilize this catalog for profile-based semantic table interpretation. 
Both example data tables can be interpreted through their data table profiles when a domain profile of the weather observation domain is provided.

\section{\tabkg{} Profile-based Semantic Table Interpretation}
\label{sec:approach}

This section describes the process of \tabkg{}'s profile-based semantic table interpretation.

\subsection{Approach Overview}

\begin{figure*}
\centering
  \includegraphics[width=0.85\textwidth]{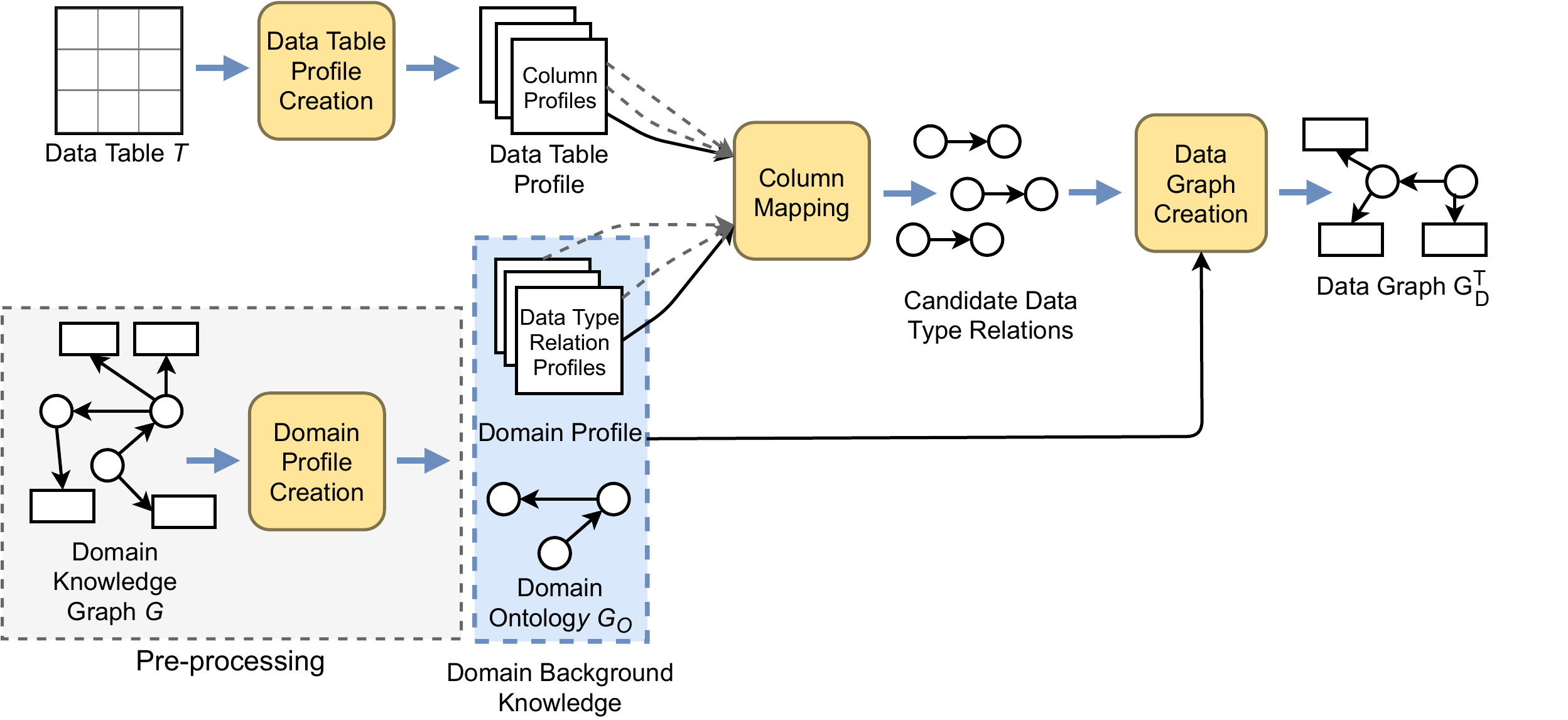}
	\caption{An overview of semantic table interpretation with \tabkg{}. Input is a data table $T$ and a domain knowledge graph $G$. The output is a data graph $G_D^T$ that represents the content of $T$ as a data graph.}
	\label{fig:pipeline}
\end{figure*}

Fig. \ref{fig:pipeline} provides an overview of our proposed \tabkg{} approach to semantic table interpretation, where a data graph $G^T_D$ is created from a data table $T$. 
To facilitate the interpretation, \tabkg{} utilizes background knowledge that includes
a domain ontology $G_{O}$ and a domain profile. 
The domain profile is generated in a pre-processing step from a domain knowledge graph $G$. 

In brief, the \tabkg{} pipeline consists of the following steps: 

\begin{enumerate}

\item \textbf{Domain Profile Creation}: In a pre-processing step, we create a domain profile from a domain knowledge graph $G$.

\item \textbf{Data Table Profile Creation}: We create a profile of the input data table $T$. 

\item \textbf{Column Mapping}: We generate candidate mappings between the columns of $T$ and the data type relations in $G_O$ using the domain profile, the data table profile, and a one-shot learning mapping function.
\item \textbf{Data Graph Creation}: We use the candidate column mappings and the domain ontology $G_O$ to create a data graph $G^T_D$ representing $T$'s content.
\end{enumerate}

In the following, we describe these steps in more detail.

\subsection{Domain Profile Creation}

The profile-based semantic table interpretation in \tabkg{} requires the availability of a domain profile. This profile can be inferred from a domain knowledge graph $G$ as described in Section \ref{sec:semantic-profiles}. The domain profile is created by computing the feature values given the literal relations in the domain knowledge graph.
This profile can be used as a lightweight domain representation. 
The domain profile can be created in a pre-processing step and become available as part of a data catalog as 
described in Section \ref{sec:semantic-domain-and-data-table-profiles}. The domain profile does not contain any entities or literals from the domain knowledge graph $G$.

\subsection{Data Table Profile Creation}

From the input data table $T$, we create a data table profile by computing the profile features based on the column values as described in Section~\ref{sec:semantic-profiles-data-table}.

\subsection{Column Mapping}
\label{sec:one-shot-learning}

With the help of the domain profile and the data table profile, we create \textit{column mappings}.

\begin{definition}
A \textbf{column mapping} is a mapping from a column $c_n$ in a data table $T$ to a data type relation $r_D \in R_D$ in the domain ontology $G_O$: $c_n \mapsto r_D$.
\end{definition}

For example, we can create a mapping from column $c_2$ of the data table illustrated in Fig. \ref{fig:example_table} to the data type relation (\voc{time}{Interval} - \voc{time}{hasBegin} - \voc{xsd}{time}) in the ontology illustrated in Fig. \ref{fig:example_schema}.

For the column mapping, we utilize a domain-independent \textit{column mapping function}. Given a column profile $\pi(c_n)$ and a data type relation profile $\pi(r_D)$, this mapping function returns a similarity score.

The column mapping function is trained in a pre-processing step. It learns from positive pairs of columns and data type relations that should be mapped (similarity: $1.0$) and negative pairs that should not (similarity: $0.0$). After training, the function predicts the similarity of previously unseen pairs.

\begin{figure*}[t]
  \centering
    \includegraphics[width=\textwidth]{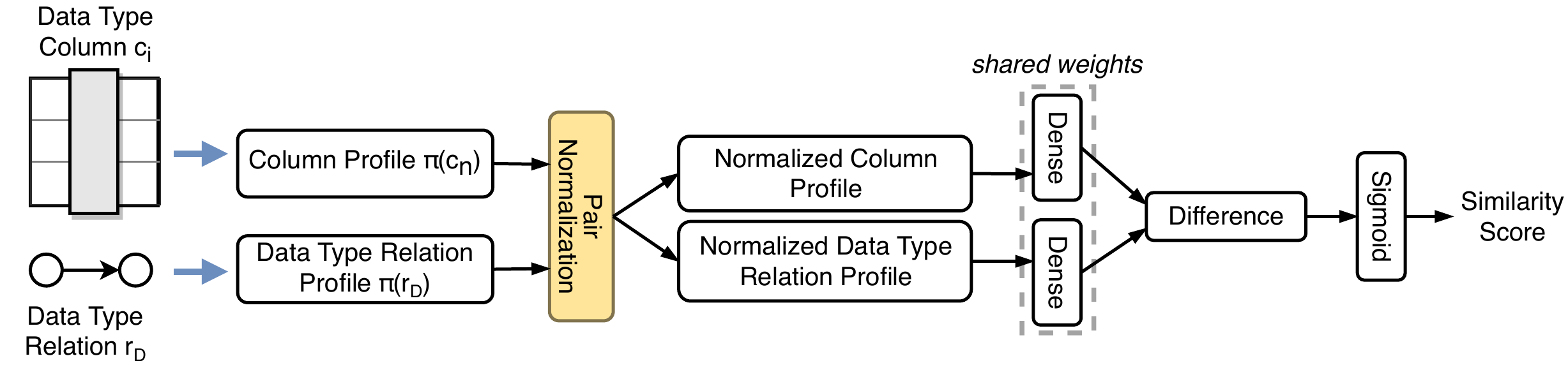}
      \caption{Architecture of the mapping function to predict the similarity between a column $c_n$ and a data type relation $r_D$.}
  \label{fig:column_mapping_model}
\end{figure*}

The architecture of the column mapping function is shown in Fig.~\ref{fig:column_mapping_model}. 
The column profile $\pi(c_n)$ and the data type relation profile $\pi(r_D)$ are taken as an input and normalized jointly. 
Then, their similarity is computed by a Siamese network that encodes both normalized profiles using the same weights and then predicts the similarity score based on the difference between the two profile encodings. As in \cite{koch2015siamese}, we use Rectified Linear Units for the hidden layers and a Sigmoid output layer. By using the Sigmoid output layer, the column mapping function learned by our Siamese network returns a similarity score in the range $[0,1]$.

Within the \tabkg{} pipeline shown in Fig.~\ref{fig:pipeline}, the column mapping function is utilized to create a set of candidate column mappings $M_{c_n}$ for each column $c_n$ in the data table $T$. In this step, we only consider mappings between columns and data type relations of the same data type (numeric, textual, temporal, spatial or Boolean).

The use of a Siamese network follows the idea used for one-shot learning for image classification \cite{koch2015siamese}. Here, the task is not only to classify images into known classes (e.g., many images showing tigers) but also to generalize towards new classes (e.g., new images showing lions). That means, the underlying classifier needs to acquire features which enable the model to successfully generalize. This is done by inducing a metric that represents the domain-independent similarity between two input feature vectors (e.g., between an unknown image and a single image showing a lion).

As we cannot train a classifier on known classes (in contrast to domain-specific approaches such as Sherlock \cite{hulsebos2019sherlock} and ColNet \cite{chen2019colnet}), we are in a one-shot learning setting as well: We may learn how to map column profiles to known data type relations. But when facing a new domain, our classifier needs to generalize towards unseen data type relations. In \tabkg{}, the similarity between a column and a data type relation is predicted based on the experience of the similarity of other profiles learned earlier.

The column profile and the data type relation profile  are normalized jointly, i.e., the features are normalized in a range between $0.0$ and $1.0$ concerning the sum of the values in both profiles. This pairwise normalization is to ensure that the normalization happens within the same range, and this range covers all the values in both sources. If both profiles were normalized in isolation, different maximum values would be both normalized to $1.0$ such that the values from different profiles were not comparable.

\subsection{Data Graph Creation}
\label{subsec:data_graph_creation}

Given a set of candidate column mappings $M_{c_n}$ with similarity scores for each column $c_n$ in the input data table $T$, we now map each table column to a data type relation in a greedy manner. First, we take the column mapping with the highest similarity score. Then, we remove all candidate column mappings with the particular column or data type relation. These two steps are repeated until all columns are mapped to a data type relation.

From the chosen column mappings set, we create the data graph $G_D^T$ that contains all data type relations resulting from the chosen mapping. $G_D^T$ needs to adhere to the following four conditions: 

\begin{enumerate}
\item The data graph covers all literal columns of $T$, and each literal column has exactly one mapping to a data type relation. 
\item The set of entities in a table row is connected via entity relations. 
\item $G_D^T$ is minimal, i.e., no relation can be removed without invalidating the previous two conditions.
\item Each class relation represented by $G_D^T$ is connected to at least one class that is part of $M_{c_n}$. 
This condition ensures semantic closeness of the data table columns and reduces the number of potential paths in the graph.
\end{enumerate}

\subsection{Creation of Training Instances for Column Mapping}

For the computation of the column mapping function, we utilize a Siamese network trained once in a pre-processing step. This training process requires the extraction of positive and negative training instances. Following Definition \ref{def:semantification} (see also Fig.~\ref{fig:pipeline}), this step requires a set of ($G$, $T$, $G_D^T$) triples, where $G$ is the domain knowledge graph, $T$ is the data table and $G_D^T$ is the resulting data graph. For each triple ($G$, $T$, $G_D^T$), each pair of data type relations in $G$ and a column in $T$ is a positive training instance. 
We select the remaining (data type relation, column) pairs from the same knowledge graph $G$ as negative instances.

In the first step, we synthetically create a set of ($G$, $T$, $G_D^T$) triples for the model training, intending to have a large dataset of positive and negative examples derived from existing data tables and knowledge graphs. Such data, i.e., a diverse set of tables and their mapping definitions to different domain ontologies, is difficult to obtain, except for manually created, task-specific research datasets \cite{PhamAKS16}, which are not large enough for training deep neural networks and do not provide enough topical and structural diversity. Therefore, we utilize existing knowledge graphs to create training data.

Given a set of knowledge graphs, we create a new dataset of triples ($G_1$, $T$, $G_2^T$). Each input knowledge graph $G$ is disjunctly (i.e., without shared triples) split into two KGs: $G_1$ and $G_2$. As $G_1$ and $G_2$ do not overlap, it is ensured that the domain profile and the data table profiles used in training are not generated from the same values. $G_1$ represents the domain knowledge graph, while $G_2$ is transformed into a data table $T$. The transformation of $G_2$ into $T$ is based on a set of domain ontology templates. A template is a directed tree with up to $k$ nodes, where $k$ is a parameter. The nodes and edges of these trees are placeholders for classes and properties. 
A set of trees is transformed into domain ontologies by replacing these placeholders with the classes and properties of $G_2$. 
From the knowledge graph $G_2$, a data graph $G_2^T$ is extracted and transformed into a data table $T$ to create the triple ($G_1$, $T$, $G_2^T$).

We aim to retrieve a set of heterogeneous data tables that represent the original knowledge graph characteristics. 
Therefore, the data table creation process incorporates several stochastic decisions in proportion to the knowledge graph statistics:

\begin{itemize}
    \item Entities and entity relations (and consequently, the literal relations) in $G$ are split at a random ratio between 25\% and 75\% into $G_1$ and $G_2$, whereas their domain ontologies remain the same.
    \item During the template creation, classes, and properties are randomly mapped to a domain ontology template, proportionally to their occurrence rate in $G$.
    \item Data type relations are added in the same manner, under the condition that each leaf node has to be connected with at least one data type relation. After adding the minimal required number of data type relations, we add data type relations as long as any of them are left and if a randomly generated number between $0$ and $1$ exceeds a predefined threshold $\delta$.
\end{itemize}

\subsection{Running Example: Data Table Creation}

\begin{figure*}
  \centering
    \includegraphics[width=\textwidth]{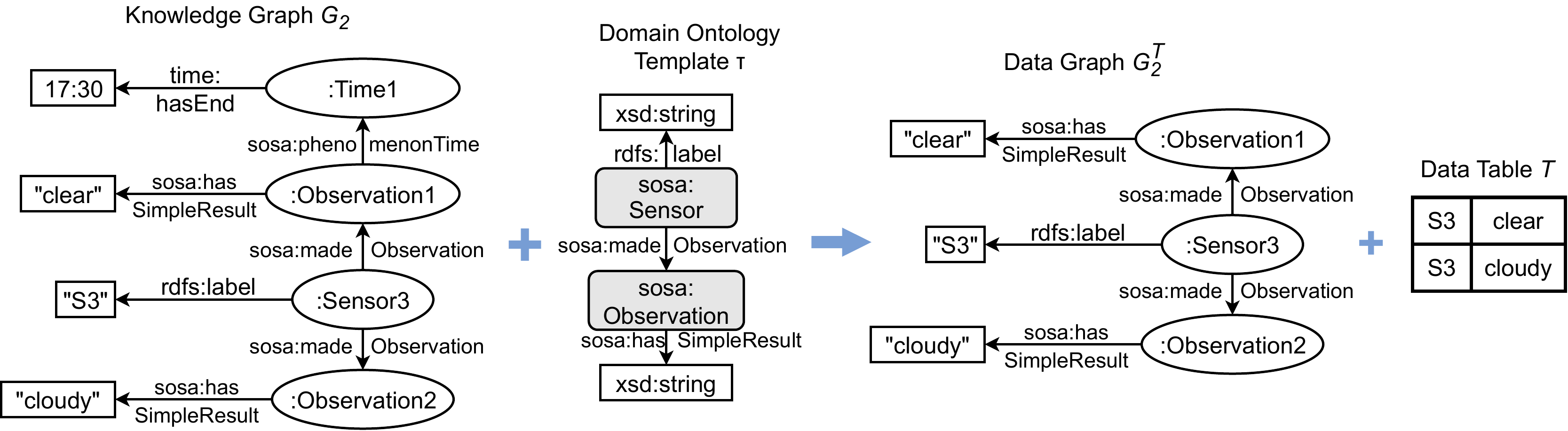}
      \caption{Creation of a data table $T$ and a data graph $G_2^T$ from a knowledge graph $G_2$ and a domain ontology template $\tau$. For brevity, we omit type relations.}
  \label{fig:table_creation}
\end{figure*}

For our running example introduced in Section \ref{sec:example}, Fig.~\ref{fig:table_creation} illustrates the transformation of a knowledge graph $G_2$ and a domain ontology template $\tau$ into a data table $T$. This transformation is performed after splitting the input knowledge graph $G$ into $G_1$ and $G_2$, where $G_1$ and $G_2$ represent different portions of $G$ (e.g., $G_1$ contains a third observation of \texttt{Sensor3}). Consequently, the triple ($G_1$, $T$, $G_2^T$) can be used as a training example where the feature weights to map $T$ to $G_2^T$ based on $G_1$'s domain profile and $T$'s data table profile are learnt.
\section{Implementation}
\label{sec:implementation}

\tabkg{} is implemented in Java 1.8. The Siamese network is trained and applied using Keras in Python 3.7. We load knowledge graphs using Apache Jena\footnote{\url{https://jena.apache.org/}}. 
We represent the column mappings inferred by \tabkg{} in the RDF Mapping Language (RML) \cite{Dimou:2014}. RML definitions are then utilized to materialize the data graph.
Data tables are provided as CSV files; knowledge graphs and data graphs as Turtle (\textit{.ttl}) files.

\subsection{Data Table Pre-Processing}
\label{sec:table-preprocessing}

Each data table interpreted in \tabkg{} runs through three pre-processing steps:

\begin{enumerate}
    \item  \textbf{Data type identification}: For each table column, we identify its data type(s) by trying to parse more than $90\%$ of the values as numeric, Boolean, spatial (Well-Known Text or Well-Known Binary format \cite{wkbwkt}) or temporal (using the dateparser library\footnote{\url{https://github.com/sisyphsu/dateparser}}). If that is not possible, the column is assigned the text data type. For the more fine-grained data types mentioned in Section \ref{sec:profiles}, we utilize regular expressions (text: URL or email), follow the algorithms proposed by Alobaid et al. \cite{alobaid2019typology} (numeric: sequential or categorical) or analyze the parsed objects (temporal: date, time, date-time; spatial: point, line string, polygon). We follow the algorithm in \cite{alobaid2019typology}, using the threshold of not more than $20$ different categories to detect categorical text values.
    \item \textbf{Key column detection}:
    When we transform a data table into a data graph based on the RML mapping, we create new entities. In RDF, each entity is identified by a Unique Resource Identifier (URI). It is important to understand how to create these URIs, as we need to re-use URIs referring to the same entity: For example, each row in our running example in Fig. \ref{fig:example_table} forms a new instance of \voc{sosa}{Observation}, together with a new URI (e.g., \voc{sosa}{Observation7}), but there should only be three different \voc{sosa}{Sensor} URIs: \voc{sosa}{SensorS1}, \voc{sosa}{SensorS2}, \voc{sosa}{SensorS3}, created using the literal values of the data type property \voc{rdfs}{label}.
    
    To create URIs, we detect data type relations representing the unique literal values of an entity as follows: (i) the data type relation is used on all instances of the class exactly once, and the literal values are unique across their instances.
    Currently, we do not consider the combinations of literal relations as identifiers \cite{heise2013scalable}; we leave such combinations for future work. 
    \item \textbf{Identifier generation}: RML transformation requires referenceable columns and instances in data tables. Therefore, we automatically generate identifiers for each column and row of the data table. If available, column names are used as part of the column identifier.
\end{enumerate}

\subsubsection{Mapping Representation in RML}

We utilize RML for storing column mappings inferred by \tabkg{} in a machine-readable format such as Turtle. The RML defines subject maps that specify how to generate subjects for each row of the data table and predicate-object maps to create predicate-object pairs. In \tabkg{}, the inferred column mappings are translated into the RML definitions according to the following four steps:

\begin{enumerate}
    \item We create one instance of \voc{rml}{source} and \voc{csvw}{Table} each, denoting relevant characteristics for parsing the data table $T$ (file location, delimiter, \ldots).
    \item For each class part of the data graph $G_D^T$, we create a new instance of \voc{rr}{TriplesMap}, together with a \voc{rr}{subjectMap} that denotes the class as well as the target node URIs.
    \item For each column mapping $c_n \mapsto r_D$, we create a \voc{rr}{predicateObjectMap} denoting the source column $c_n$, the data type property and a reference to the data type relation $r_D$.
    \item For each class relation $r \in R_{OC}$ in the domain ontology $G_O$, we create a \voc{rr}{predicateObjectMap} connecting the respective entities and the object property.
\end{enumerate}

\subsubsection{Running Example: RML Mapping Definitions}

Listing \ref{lst:rml} in the Appendix provides an example of the RML definitions that were automatically generated for our running example introduced in Section \ref{sec:example} -- without the time intervals, for brevity. Instances of \voc{sosa}{Sensor} and \voc{sosa}{Observation} are created alongside their relations. The sensor labels in the third column were detected as identifiers, i.e., we create node URIs as \textit{https://www.w3.org/TR/vocab-ssn/\allowbreak{}Sensor\{col3\}}\footnote{The RML template definitions do not allow to use prefixes.}.

Listing \ref{lst:rml2} in the Appendix provides the resulting Turtle file representing the knowledge graph inferred from the input data table. The correct mapping of the third line shown in Fig. \ref{fig:correct_mapping_example} is contained here.

\section{Evaluation Setup}
\label{sec:evaluation}

The goal of the evaluation is to assess the performance of \tabkg{} 
concerning the semantic table interpretation effectiveness using lightweight semantic profiles. 
%


In this section, we will first describe how the data tables and domain ontologies are selected, how data table profiles and domain profiles are built and which baselines we compare to.

We make the scripts to extract the evaluation datasets and the code for training the Siamese network publicly available\footnote{\url{https://github.com/sgottsch/Tab2KG}}.

\subsection{Domain Ontologies and Vocabularies}

In the real-world applications of \tabkg{}, we require a domain ontology and a dataset that is modelled using this ontology to build the domain profile.

In our experiments, we aim to assess the performance of \tabkg{} in different domains for mapping real-world tables. This evaluation requires annotated datasets which provide the tables and their ontology mappings. 
In the existing datasets such as SemTab~\cite{jimenez2020semtab}, the tables are typically annotated using cross-domain ontologies such as DBpedia and schema.org. 
To facilitate evaluation, we extract domain ontologies from tables within specific domains. 
This extraction results in the domain-specific parts of the existing cross-domain ontologies.

For the ontology extraction, we use two approaches: a \textit{pairwise setting} ($P$) that uses one table and a \textit{set-based setting} ($S$) that uses a union of tables to represent a domain. 
These approaches are discussed in Section~\ref{subsec:pairwise_setbased}.

Statistics of the domain ontologies used in the evaluation and examples of their involved classes  are presented in Table~\ref{tab:ontologies}.

\begin{table*}
\centering
\caption{Domain ontologies used during evaluation, together with their used vocabulary, the number of classes and literal relations. In total, \tabkg{} is evaluated against 34 domain ontologies. ST and SE use the same domain ontologies which are grouped together as ST$_S$ and SE$_S$, respectively.}
\label{tab:ontologies}
\setlength{\tabcolsep}{3pt}
\footnotesize
\begin{tabular}{@{}llrrl@{}}
\toprule
\textbf{Domain} & \textbf{Vocabulary} & \textbf{\#Classes} & \textbf{\#Relations} & \textbf{Classes} \\ \midrule
\textbf{So$_S$}: Soccer & schema.org & 4 & 21 & Player, League, SportsClub, SportsTeam \\
\textbf{WA$_S$}: \begin{tabular}[c]{@{}l@{}}Weapon ads.\end{tabular} & schema.org & 7 & 37 & PersonOrOrganization, Place, PostalAddress, Firearm, Offer, \ldots \\
\textbf{ST$_S$}, \textbf{SE$_S$}: City & DBpedia & 15 & 55 & EducationalInstitution, BasketballPlayer, BasketballTeam, Settlement, \ldots \\
\textbf{ST$_S$}, \textbf{SE$_S$}: University & DBpedia & 14 & 52 & City, OfficeHolder, Person, Country, Artist, Organisation, ChristianBishop, \ldots \\
\textbf{ST$_S$}, \textbf{SE$_S$}: Governor & DBpedia & 9 & 23 & City, Person, Country, University, Senator, OfficeHolder, Town, Region \\
\textbf{ST$_S$}, \textbf{SE$_S$}: Building & DBpedia & 9 & 25 & Architect, GovernmentAgency, City, Country, Region, Organisation, \ldots \\
\textbf{ST$_S$}, \textbf{SE$_S$}: Song & DBpedia & 8 & 21 &  MusicalArtist, Band, Single, Company, City, Settlement, PopulatedPlace \\
\textbf{ST$_S$}, \textbf{SE$_S$}: Play & DBpedia & 6 & 15 & Person, Language, Writer, MusicalArtist, Album \\
\multicolumn{5}{c}{\vdots} \\ \bottomrule
\end{tabular}
\end{table*}

\subsection{Datasets}
\label{sec:datasets}

The Siamese network training requires a dataset that spans multiple vocabularies and domains to ensure generalization. 
However, the domain ontologies shown in Table~\ref{tab:ontologies} are based on schema.org and the DBpedia ontology only. To feed and evaluate the Siamese network with data from a more diverse set of vocabularies, we created a new synthetic dataset automatically extracted from GitHub repositories dealing with knowledge graphs. The column mapping function trained on the train split of this dataset is expected to generalise to other domains and vocabularies represented in our datasets. Therefore, we use this mapping function for all datasets in the evaluation and do not train any dataset-specific column mapping functions.

\subsubsection{Synthetic GitHub Dataset}

The GitHub advanced code search\footnote{\url{https://github.com/search/advanced}} provides access to millions of data repositories. To collect knowledge graphs from GitHub, we selected files larger than 5 KB with the specific file extensions\footnote{\texttt{ttl}, \texttt{rdf}, \texttt{nt}, \texttt{nq}, \texttt{trix}, \texttt{n3}, \texttt{owl}} that contain the text ``\texttt{xsd}'' or ``\texttt{XSDSchema}''. To ensure the heterogeneity of our dataset, we limited the number of files per GitHub repository to three. Each file that was successfully parsed as a knowledge graph with more than $50$ statements including at least $25$ literal relations was added to our dataset. This way, we obtained $3,922$ files.

We set the parameter for maximum tree size $k=3$, and the parameter for adding data type relations $\delta=0.2$. The knowledge graphs set was split into a training set ($90\%$) and a test set ($10\%$). Knowledge graphs from the same repository were not included in the same set.

\subsubsection{Test Datasets}
\label{subsubsec:test_datasets}

We use the following datasets for evaluating our approach:

\begin{itemize}
    \item \textbf{GitHub (GH):} The test split of the synthetic GitHub dataset, without a restriction on the used vocabularies.
    \item \textbf{Soccer (So):} $12$ data tables regarding soccer players and their teams, annotated with the \textit{schema.org} vocabulary \cite{PhamAKS16}.
    \item \textbf{Weapon Ads (WA):} $15$ data tables about weapon advertisements, annotated with the \textit{schema.org} vocabulary \cite{taheriyan2016leveraging}.
    \item \textbf{SemTab (ST):} A collection of $2,281$ data tables extracted from the T2Dv2 web table corpus, Wikipedia and others, annotated with the DBpedia ontology \cite{jimenez2020semtab}.
    \item \textbf{SemTab Easy (SE):} A subset of ST, including those $676$ data tables whose columns are mapped to one class only. Only classes appearing in the \tkmatch{} corpus \cite{ritze2015matching} are included.
\end{itemize}

For all data tables contained in these datasets, we set the following constraints:
\begin{enumerate}
    \item The input table file is parseable as a CSV file without errors.
    \item There are no classes that are instantiated multiple times in the same row. This condition is to avoid cyclic structures. We discuss limitations regarding cyclic structures in Section \ref{subsec:limitations}.
    \item There is no pair of columns with identical values in the same table. This condition is to avoid randomness during the evaluation.
    \item The table has at least two columns which are mapped to a data type relation. This condition is to ensure that the semantic table interpretation task does not become trivial in such cases.
\end{enumerate}

Following the process to create data tables via domain ontology templates described in Section~\ref{subsec:data_graph_creation}, the GH data tables did not violate these conditions. Also, these conditions were not violated by the So and WA data tables. In the case of the SemTab data tables, we had to skip $3,017$ non-parseable data tables (e.g., because of column values containing non-escaped delimiters such as in \textit{"Bret "Hitman" Hart"}, $2,233$ cyclic data tables (e.g., mathematicians and their supervisors), $520$ data tables with identical column values (e.g., columns mapped to \voc{dbo}{birthDate} and \voc{dbo}{birthYear} but with identical values) and $6,179$ data tables with one column only.

\subsection{Profile \& Test Data Generation}
\label{subsec:pairwise_setbased}

It is important to emphasize the difference in the evaluation setting compared to typical evaluation using the previously mentioned datasets such as ST: In the experiments conducted by \cite{PhamAKS16, chen2020linkingpark, zhang2017effective}, target general-purpose knowledge graphs such as DBpedia or Wikidata are given. Each data table in the test set is then mapped to the nodes in such a knowledge graph. In our evaluation setting, we assume that no data instances are given, i.e., an instance lookup is not possible. 

\begin{figure*}
  \centering
    \includegraphics[width=0.75\textwidth]{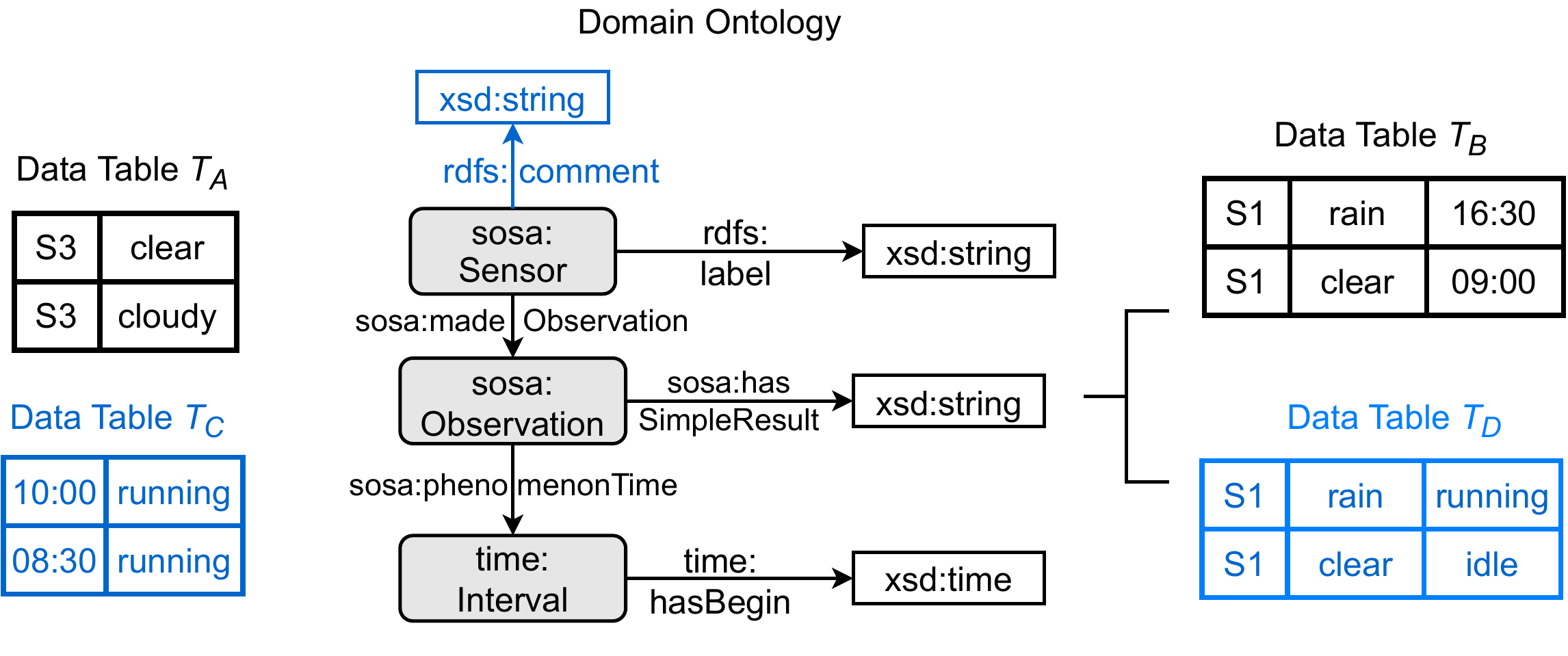}
      \caption{Comparison of the pairwise (black) and the set-based setting (black and blue) for creating training and test instances. In the pairwise setting, one data table ($T_B$) represents the domain knowledge of the other data table ($T_A$). In the domain ontology approach, we first create a domain ontology from a set of data tables ($T_B$ and $T_D$) and then interpret several data tables ($T_A$ and $T_C$) against the domain ontology each. In this example, ``running'' and ``idle'' are comments (\voc{rdfs}{comment}) about specific sensors.}
  \label{fig:pairwise_vs}
\end{figure*}

Instead, a domain profile and a domain ontology are provided which are derived via the pairwise and the set-based setting. Fig.~\ref{fig:pairwise_vs} exemplifies these two approaches which are described in more detail in the following. While we consider pairs of data tables in the pairwise setting, we create a domain ontology based on a set of tables in the set-based setting.

The set-based setting and the pairwise setting reflect different application scenarios of \tabkg{}. 
The pairwise setting considers smaller domain ontologies, where the directly corresponding concepts are available in the ontology and a table. An example task for the pairwise setting is table augmentation, where a given table is populated with additional rows, columns or cell values~\cite{zhang2020web}. The set-based setting reflects traditional semantic table interpretation and involves larger domain ontologies, with concepts potentially represented in several tables.

\subsubsection{Pairwise Setting}

In the pairwise setting, we select data table pairs, such that one data table mimics the domain knowledge graph, from which we can derive a domain profile.

Following our setting defined in Definition \ref{def:semantification} and illustrated in Fig.~\ref{fig:pipeline}, the datasets are transformed into a set of triples ($G$, $T$, $G_D^T$), consisting of a data table $T$, a domain knowledge graph $G$ and the mapping definition which transforms data table $T$ into the data graph $G_D^T$. Technically, we extract a set of test instances, consisting of (i) a \textit{.ttl} file representing the domain knowledge graph $G$, and (ii) a \textit{.csv} file representing the data table $T$ and a \textit{.rml} file representing the mapping from $T$ to $G$'s domain ontology. 
To transform the datasets into such test instances, we identify pairs of data tables where the columns of the first data table are a subset of the second data table's columns (i.e., the columns of the first data table are all mapped to data type relations which also exist in the column mapping of the second data table). Then, the second data table represents the domain knowledge. 
Table \ref{tab:datasets} provides an overview of the datasets used during training and evaluation under these conditions. ST and SE cover all data tables from the domain ontologies illustrated in Table~\ref{tab:ontologies}.

\setlength{\tabcolsep}{0.5em}
\begin{table*}
\centering
\caption{Datasets used in the evaluation based on the pairwise setting. \# is the number of ($G$, $T$, $G_D^T$) triples that we generated. The other columns contain average values. For example, the tables ($T$) in the ST dataset have about four columns on average, while their paired domain knowledge graphs have $5.5$ data type relations on average. Following the pairwise extraction process, the number of data type relations in the domain knowledge graphs is always equal or higher than the number of columns in the paired tables.}
\label{tab:datasets}
\footnotesize{}
\begin{tabular}{lr|r|rr}\toprule
 &  & \multicolumn{1}{c|}{\textbf{Tables}} & \multicolumn{2}{l}{\textbf{Domain Knowledge Graphs}} \\ \midrule
\multicolumn{1}{c}{\textbf{Dataset}} & \multicolumn{1}{c|}{\textbf{\begin{tabular}[c]{@{}c@{}}\#\end{tabular}}} & \multicolumn{1}{c|}{\textbf{Columns}} & \multicolumn{1}{c}{\textbf{\begin{tabular}[c]{@{}c@{}}Data Type\\ Relations\end{tabular}}} & \multicolumn{1}{c}{\textbf{\begin{tabular}[c]{@{}c@{}}Class\\Relations\end{tabular}}} \\ \midrule
GitHub$_P$ (Training) & $8,078$ & $2.18$ & $3.03$ & $1.35$ \\ \midrule
\textbf{GH$_P$}: GitHub (Test) & $891$ & $2.24$ & $3.16$ & $1.44$ \\
\textbf{So$_P$}: Soccer & $15$ & $6.25$ & $9.38$ & $2.75$ \\
\textbf{WA$_P$}: Weapon-Ads & $16$ & $10.57$ & $11.2$ & $4.4$ \\
\textbf{ST$_P$}: SemTab & $233$ & $4.09$ & $5.54$ & $1.26$ \\
\textbf{SE$_P$}: SemTab Easy & $125$ & $4.2$ & $5.53$ & $0.0$ \\ \bottomrule
\end{tabular}
\end{table*}

\subsubsection{Set-based Setting}

In the set-based setting, we create a set of domain ontologies, each based on a set of data tables. For So and WA, we create one domain ontology each, based on all available data tables. In the case of ST and SE, we create several domain ontologies by grouping together data tables based on the class which most columns are mapped to. We only consider domain ontologies from at least 10 data tables and with at least five data type relations. Due to the diversity of domains in the GitHub data set, we do not evaluate GH with the set-based setting.

Table~\ref{tab:ontologies} provides an overview of the domain ontologies used in our evaluation.

\subsection{Baselines}

We compare \tabkg{} against the following three semantic table interpretation baselines:

\begin{itemize}
    \item \textbf{\dsl{}}: The Domain-independent Semantic Labeler \cite{PhamAKS16} uses logistic regression on a set of hand-crafted features; some of them compare value pairs at the instance-level. It has been shown to outperform previous approaches such as the SemanticTyper \cite{ramnandan2015assigning}. We train \dsl{} on the same GitHub training data set as \tabkg{}.
    \item \textbf{\dslless{}}: The Domain-independent Semantic Labeler without using value similarity at the instance level. In contrast to \tabkg{} and the other baselines, \dsl{} utilizes a domain-specific data graph. As \tabkg{} is solely based on the domain profile, \dsl{} is in an advantageous setting that does not entirely reflect our setting. Therefore, we remove features at the instance-level for the \dslless{} baseline.
    \item \textbf{\tkmatch{}}: \tkmatch{} \cite{ritze2015matching} performs semantic table interpretation on the instance level. In contrast to other approaches \cite{cremaschi2020fully, MTab, zhang2017effective} that rely on a costly knowledge graph lookup at runtime, \tkmatch{} creates an index over the instances of frequently used DBpedia classes and is thus commonly used as a baseline for semantic table interpretation approaches \cite{zhang2020novel, EfthymiouHRC17,chen2019colnet}. It combines a ranking of entities found in the lookup phase for column type identification and data type-specific similarity measures (Levenshtein distance for strings, deviation for numbers, and deviation of years for dates) for property identification. \tkmatch{} assumes that a table only describes one entity class at a time and thus does not consider class relations.
\end{itemize}

\section{Evaluation Results}
\label{sec:evaluation_results}

In this section, we describe the training performance and the evaluation results based on the evaluation setup described before.

\subsection{Accuracy of the Column Mapping Function}
\label{subsec:acc_column_mapping}

We train our Siamese network on the GitHub training dataset for $1,000$ epochs, a batch rate of $100$ and a learning rate of $0.0001$, following Hsiao et al.'s approach for one-shot image classification \cite{hsiao2019malware}. We apply an early stopping criterion if the validation accuracy has not improved within 100 epochs. We use $256$ dimensions for the hidden layer. $10\%$ of the training dataset are used for validation. The feature vectors contain histograms with $10$ buckets.

After training for all epochs on the synthetic training dataset, the Siamese network has an accuracy of $0.959$ and a loss (binary cross-entropy) of $0.149$ on the validation set. On the test set, it achieves an accuracy of $0.952$, when treating scores of greater than $0.5$ as candidates for column mapping.

We experimented with different train/test splits (90/10 and 80/20). We achieve a similar accuracy on the splits, namely 0.948 on the 80/20 and 0.952 on the 90/10 split. Whereas slight variations are expected, we do not observe any remarkable drop in performance if less training data is provided to the algorithm.

\subsubsection{Hyperparameter Tuning}

To identify the best combination of hyperparameters, we performed a grid search over the number of hidden layers and their dimensions. The results with respect to the classification accuracy are shown in Table~\ref{tab:grid_search}. Following these results, using one hidden layer is sufficient, and we set its dimension to $256$.

\begin{table}[ht]
\centering
\caption{Grid search results (accuracy on the test set) over the number of hidden layers and their dimension. Higher accuracy values are marked darker.}
\label{tab:grid_search}
\setlength{\tabcolsep}{2pt}
        \begin{tabular}{ccRRRRRR}
        \toprule
        & & \multicolumn{6}{c}{\textbf{Hidden Layers Dimension}} \\
        & & \multicolumn{1}{c}{\textbf{16}} & \multicolumn{1}{c}{\textbf{32}} & \multicolumn{1}{c}{\textbf{64}} & \multicolumn{1}{c}{\textbf{128}} & \multicolumn{1}{c}{\textbf{256}} & \multicolumn{1}{c}{\textbf{512}} \\ \midrule
      \multirow{3}{*}{\textbf{\begin{tabular}[c]{@{}c@{}}\# Hidden\\ Layers\end{tabular}}}  & \textbf{1} & 0.947 & 0.943 & 0.949 & 0.950 & 0.952 & 0.937 \\
      &   \textbf{2} & 0.935 & 0.942 & 0.946 & 0.948 & 0.939 & 0.907 \\
      &   \textbf{3} & 0.891 & 0.930 & 0.909 & 0.934 & 0.939 & 0.821 \\ \bottomrule
        \end{tabular}
\end{table}

\subsubsection{Ablation study}

To determine the effect of the selected features on the model's performance, we have performed an ablation study where we trained the Siamese network after removal of specific feature groups as listed in Section~\ref{sec:profile-features}. Table~\ref{tab:ablation} shows the model's classification performance on the GH test set. As expected, the combination of all selected features into the profiles leads to the highest accuracy. Among the different feature groups, the removal of the basic statistics features leads to the highest drop in performance. 
In general, the ablation study demonstrates that all features contribute to increase the accuracy of \tabkg{}.

\begin{table}[t]
\centering
\caption{Ablation study: Number of true positives (TP), false positives (FP), true negatives (TN), false negatives (FN) and accuracy (Acc.) on the test set.}
\label{tab:ablation}
\setlength{\tabcolsep}{3pt}
\begin{tabular}{@{}lrrrrr@{}}
\toprule
 & \multicolumn{1}{c}{\textbf{TP}} & \multicolumn{1}{c}{\textbf{FP}} & \multicolumn{1}{c}{\textbf{TN}} & \multicolumn{1}{c}{\textbf{FN}} & \multicolumn{1}{c}{\textbf{Acc.}} \\ \midrule
\textbf{All features} & 1,873 & 125 & 1,810 & 62 & \textbf{0.952} \\
\textbf{No data type features} & 1,895 & 251 & 1,684 & 40 & 0.924 \\
\textbf{No completeness features} & 1,865 & 141 & 1,794 & 70 & 0.945 \\
\textbf{No basic statistical features} & 1,822 & 194 & 1,741 & 113 & 0.921 \\
\textbf{No histograms and quantiles} & 1,862 & 132 & 1,803 & 73 & 0.947 \\ \bottomrule
\end{tabular}
\end{table}


\subsection{Semantic Table Interpretation Results}
\label{subsec:stir}

Until now, we evaluated the accuracy of the column mapping function, i.e. \tabkg{}'s ability to provide correct data type relation mappings. Now, we evaluate the entire semantic table interpretation process. While we expect a similar accuracy for data type relation mappings for the GitHub dataset, the results depend on the accuracy of class relations mappings and can vary across datasets. The evaluation in this section follows three steps: we measure accuracy (i) of the whole semantic table interpretation, (ii) of the column mapping and graph creation in isolation, and (iii) of the semantic table interpretation on different domains.

\subsubsection{Accuracy of the Semantic Table Interpretation}

\setlength{\tabcolsep}{0.5em}
\begin{table*}[t]
\centering
\caption{Semantic table interpretation performance of \tabkg{}, compared to the baselines on five datasets. We report the accuracy, i.e. the percentage of correctly identified data type relations ($R_{OD}$) and class relations ($R_{OC}$) in the datasets. Averages are computed in relation to the number of class relations plus data type relations in the datasets. T2KMatch is only evaluated on SE$_P$ and SE$_S$, as it assumes one entity class per table only.
}
\label{tab:evaluation-total}
\begin{tabular}{lrrrrrrrrrrrrr}\toprule
 & \multicolumn{1}{c}{\textbf{GH$_P$}} & \multicolumn{1}{c}{\textbf{So$_P$}} & \multicolumn{1}{c}{\textbf{So$_S$}} & \multicolumn{1}{c}{\textbf{WA$_P$}} & \multicolumn{1}{c}{\textbf{WA$_S$}} & \multicolumn{1}{c}{\textbf{ST$_P$}} & \multicolumn{1}{c}{\textbf{ST$_S$}} & \multicolumn{1}{c}{\textbf{SE$_P$}} & \multicolumn{1}{c}{\textbf{SE$_S$}} & \multicolumn{1}{c}{\textbf{Average$_P$}} & \multicolumn{1}{c}{\textbf{Average$_S$}} & \multicolumn{1}{c}{\textbf{Average}} \\ \hline
\textbf{\dsl{}}     & $0.89$ & $0.56$ & $0.52$ & $0.40$ & $0.39$ & $0.75$ & $0.51$ & $0.81$ & $0.56$ & $0.79$ & $0.52$ & $0.62$ \\ 
\textbf{\dslless{}} & $0.87$ & $0.45$ & $0.46$ & $0.44$ & $0.36$ & $0.68$ & $0.52$ & $0.75$ & $0.59$ & $0.76$ & $0.54$ & $0.61$ \\ 
\textbf{\tkmatch{}} & - & - & - & - & - & - & - & $0.41$ & $0.35$ & $0.41$ & $0.35$ & $0.36$ \\ \midrule
\textbf{\tabkg{}} & $0.89$ & $0.69$ & $0.54$ & $0.49$ & $0.43$ & $0.81$ & $0.66$ & $0.84$ & $0.69$ & \bm{$0.84$} & \bm{$0.66$} & \bm{$0.72$} \\ \bottomrule
\end{tabular}
\end{table*}

We evaluate the performance of the semantic table interpretation achieved by \tabkg{} compared to the baselines.
Table \ref{tab:evaluation-total} shows how the approaches perform on the different datasets, measured using accuracy, i.e., the percentage of the columns correctly mapped to data type relations and correctly identified class relations.
We do not evaluate the performance of \tkmatch{} on other datasets than SE, as \tkmatch{} assumes one entity class per table only.

As we can observe in Table \ref{tab:evaluation-total}, the accuracy of the approaches varies considerably across the datasets, which can be explained by the different domain ontology and dataset characteristics shown in Table \ref{tab:ontologies} and Table~\ref{tab:datasets}.

All methods perform worst on the WA dataset, which has the most columns, data type relations and class relations within the datasets and domain ontologies. This result unsurprisingly demonstrates that semantic table interpretation gets more difficult the more columns and relations are involved. 
In all cases except for the GH dataset, where \tabkg{} and \dsl{} show similar performance, \tabkg{} 
achieves higher accuracy than the baselines concerning column mapping. 
Even though \tabkg{} utilizes less information than \dsl{}, \tabkg{} performs better by $10$ percentage points on average on this task.

The semantic interpretation of tables with domain ontology that were created with the set-based setting is more difficult than with the pairwise setting (\tabkg{} accuracy: $0.84$ vs. $0.66$). While the domain ontologies of the set-based setting provide richer domain knowledge, the increased number of classes and relations, i.e., the increased number of candidate mappings, leads to more misinterpretations.

Surprisingly, \dsl{} is also outperformed by \dslless{} in three cases (WA$_P$, ST$_S$, SE$_S$). To explain this behavior, we have computed the percentage of table values that also appear in the mapped data table relations: GH$_P$ ($33.38\%$), So$_P$ ($16.92\%$), ST$_P$ ($14.67\%$), SE$_P$ ($10.63\%$), WA$_P$ ($2.65\%$). This observation shows that profile-based semantic table interpretation can outperform instance-level approaches when the overlap between the data table and the instances in the domain knowledge graph is low.

\begin{table*}
\centering
\caption{Semantic table interpretation performance of \tabkg{} in detail, compared to the baselines on five datasets created in the pairwise setting. The results are reported as the accuracy of class relations ($R_{OC}$) and data type relations ($R_{OD}$). Averages are computed in relation to the number of class relations and data type relations in the datasets, respectively. T2KMatch is only evaluated on SE$_P$, as it assumes one entity class per table only.}
\label{tab:eval_split_p}
\begin{tabular}{@{}lrrrrrrrrrrrrr@{}}
\toprule
 & \multicolumn{2}{c}{\textbf{GH$_P$}} & \multicolumn{2}{c}{\textbf{So$_P$}} & \multicolumn{2}{c}{\textbf{WA$_P$}} & \multicolumn{2}{c}{\textbf{ST$_P$}} & \multicolumn{1}{c}{\textbf{SE$_P$}} & \multicolumn{2}{c}{\textbf{Average$_P$}} \\ \midrule
 & \multicolumn{1}{c}{\textbf{$R_{OC}$}} & \multicolumn{1}{c}{\textbf{$R_{OD}$}} & \multicolumn{1}{c}{\textbf{$R_{OC}$}} & \multicolumn{1}{c}{\textbf{$R_{OD}$}} & \multicolumn{1}{c}{\textbf{$R_{OC}$}} & \multicolumn{1}{c}{\textbf{$R_{OD}$}} & \multicolumn{1}{c}{\textbf{$R_{OC}$}} & \multicolumn{1}{c}{\textbf{$R_{OD}$}} & \multicolumn{1}{c}{\textbf{$R_{OD}$}} & \multicolumn{1}{c}{\textbf{$R_{OC}$}} & \multicolumn{1}{c}{\textbf{$R_{OD}$}}\\
 \midrule
\textbf{\dsl{}} & $0.62$ &$0.93$ & $0.69$ & $0.52$ & $0.44$ & $0.38$ & $0.88$ & $0.73$ & $0.81$ & $0.64$ & $0.83$  \\
\textbf{\dslless{}} &$0.50$ & $0.94$ & $0.50$ & $0.43$ & $0.42$ & $0.45$ & $0.72$ & $0.68$ & $0.75$ & $0.52$ & $0.81$ \\
\textbf{\tkmatch{}} & - & - & - & - & - & - & - & - & $0.41$ & - & $0.41$ &  \\ \midrule
\textbf{\tabkg{}} & $0.73$ & $0.94$ & $0.73$ & $0.68$ & $0.23$ & $0.61$ & $0.78$ & $0.81$ & $0.84$ & \bm{$0.69$} & \bm{$0.87$} \\ \bottomrule
\end{tabular}
\end{table*}

\begin{table*}
\centering
\caption{Semantic table interpretation performance of \tabkg{} in detail, compared to the baselines on four datasets created using the set-based setting. The results are as the accuracy of class relations ($R_{OC}$) and data type relations ($R_{OD}$). Averages are computed in relation to the number of class relations and data type relations in the datasets, respectively. T2KMatch is only evaluated on SE$_S$, as it assumes one entity class per table only.}
\label{tab:eval_split_s}
\begin{tabular}{@{}lrrrrrrrrrrr@{}}
\toprule
 & \multicolumn{2}{c}{\textbf{So$_S$}} & \multicolumn{2}{c}{\textbf{WA$_S$}} & \multicolumn{2}{c}{\textbf{ST$_S$}} & \multicolumn{1}{c}{\textbf{SE$_S$}} & \multicolumn{2}{c}{\textbf{Average$_S$}} \\ \midrule
 & \multicolumn{1}{c}{\textbf{$R_{OC}$}} & \multicolumn{1}{c}{\textbf{$R_{OD}$}} & \multicolumn{1}{c}{\textbf{$R_{OC}$}} & \multicolumn{1}{c}{\textbf{$R_{OD}$}} & \multicolumn{1}{c}{\textbf{$R_{OC}$}} & \multicolumn{1}{c}{\textbf{$R_{OD}$}} & \multicolumn{1}{c}{\textbf{$R_{OD}$}} & \multicolumn{1}{c}{\textbf{$R_{OC}$}} & \multicolumn{1}{c}{\textbf{$R_{OD}$}}\\
 \midrule
\textbf{\dsl{}} & $0.96$ & $0.39$ & $0.51$ & $0.34$ & $0.32$ & $0.54$ & $0.56$ & $0.36$ & $0.54$  \\
\textbf{\dslless{}} & $0.78$ & $0.37$  & $0.49$ & $0.30$  & $0.31$ & $0.55$& $0.59$ & $0.35$ & $0.54$ \\
\textbf{\tkmatch{}} & - & - & - & - & - & - & $0.35$ & - & $0.35$ &  \\ \midrule
\textbf{\tabkg{}} & $0.70$ & $0.49$ & $0.40$ & $0.43$ & $0.54$ & $0.68$ & $0.69$ & \bm{$0.53$} & \bm{$0.67$} \\ \bottomrule
\end{tabular}
\end{table*}

\subsubsection{Accuracy of the Column Mapping and Graph Creation}

Now, we assess the results of the column mapping (percentage of correctly mapped columns to the data type relations $R_{OD}$) and the graph creation (correctly identified class relations $R_{OC}$) in isolation. We report the results achieved by \tabkg{} in comparison to the baselines in Table \ref{tab:eval_split_p} (pairwise) and in Table~\ref{tab:eval_split_s} (set-based). Regarding column mapping in the pairwise setting, \tabkg{} performs best on average, outperforming \dsl{} by $4$ percentage points. A similar distribution of results, but with less accuracy, can be observed in the set-based setting.

The high accuracy of $0.95$ for the column mapping on the GitHub test dataset reported in Section~\ref{subsec:acc_column_mapping} is confirmed in Table~\ref{tab:eval_split_p}. Accuracy drops slightly to $0.94$ which can be explained by the greedy process described in Section~\ref{subsec:data_graph_creation}: during the data graph creation, the column mappings are not decided in isolation: an erroneous selection of a column mapping influences the subsequent selections.

In general, the class relation mapping results in less accuracy than the column mapping. One reason is that errors propagate along the pipeline, i.e., a wrongly mapped data type relation invokes an erroneous class relation mapping.

\subsubsection{Accuracy on Different Domains}
ST$_S$ and SE$_S$ cover various domains as shown in Table~\ref{tab:ontologies}. We evaluated the accuracy of \tabkg{} on ST$_S$ regarding the different domains. For the domains with a large number of classes and relations shown in Table~\ref{tab:ontologies}, the following accuracy is achieved: Play ($0.70$), Building ($0.68$), Song ($0.55$), City ($0.49$), University ($0.37$) and Governor ($0.35$). Three domains with a lower number of classes and relations, namely Airport, Scientist and Ligament, achieve an accuracy of $1.0$.

\subsection{Error Analysis}

By inspection of the results, we have identified two typical sources of erroneous results in \tabkg{}:
\begin{itemize}
\item Value formatting: For example, the soccer dataset has data tables with column values such as ``Germany'', whereas the domain knowledge graphs had ``GER'' as a country label. Thus, the respective profile features were highly different. Regarding high-quality domain knowledge graphs that, for example, distinguish between labels and abbreviations, the error rate should be lower.
\item Data type differences: For example, the SemTab dataset has elevations of mountains both denoted via integer values (``5291'') and as text (``2,858 ft (871 m)''). Again, the proper use of an ontology and its \voc{rdfs}{range} constraints on properties should alleviate this problem.
\end{itemize}

Overall, our evaluation results demonstrate that domain profiles in combination with one-shot learning adopted by \tabkg{} are an effective method for semantic table interpretation. This method does not require any instance lookup and achieves the highest accuracy on several datasets compared to the baselines.

\section{Discussion}
\label{sec:discussion}

In this article, we presented \tabkg{} -- an approach for tabular data semantification. 
\tabkg{} relies on domain profiles that enrich the relations in a domain ontology and serve as a lightweight domain representation. 
\tabkg{} matches these profiles with tabular data using one-shot learning. Our evaluation shows that \tabkg{} outperforms the baselines for semantic table interpretation of five real-world datasets. 
In future work, we plan to consider integrating user feedback into the \tabkg{} pipeline to support an extension of the domain ontology in cases where tabular data contains previously unseen relations. 

The representation of data tables as data graphs opens up several possibilities for making the lives of data scientists easier -- with benefits including increased efficiency and robustness of data analytics workflows. Automating semantic table interpretation, \tabkg{} takes an essential step in assisting domain experts and adds an important layer of abstraction to DAWs.

\subsection{Limitations}
\label{subsec:limitations}

We identified few limitations of \tabkg{}, which can be attributed to the idea of using semantic lightweight dataset profiles, without requiring knowledge about particular data instances.

\subsubsection{Column Mapping without Knowing the Dataset Instances}

\begin{figure}[ht]
    \centering
    \includegraphics[width=0.85\linewidth]{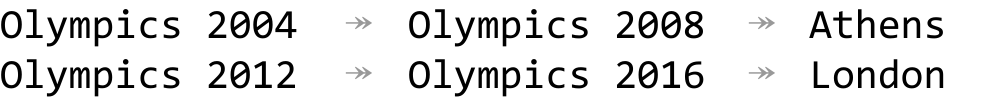}
    \caption{Cyclic class relation: Did the London Olympic Games happen in 2012 or in 2016?}
    \label{fig:limitation02}
\end{figure}

\begin{figure}[ht]
    \centering
    \includegraphics[width=0.85\linewidth]{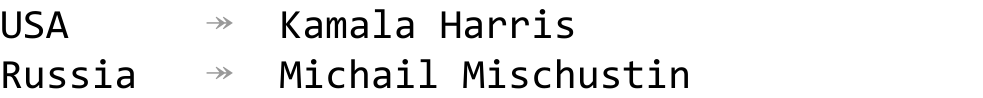}
    \caption{Class relations connecting the same classes: Is Kamala Harris the president or the vice president of the US?}
    \label{fig:limitation01}
\end{figure}

In contrast to approaches that perform semantic table interpretation at the instance level, i.e., with the help of the instance lookup in the domain knowledge graph, \tabkg{} derives column mappings from statistical features in the domain profile. We have identified two limitations to this approach:

\begin{itemize}
\item Cyclic class relations: Currently, we do not address cyclic relations in the domain ontology, as for example (\voc{dbo\footnote{\url{http://dbpedia.org/ontology/}}}{Event} \voc{dbo}{nextEvent} \voc{dbo}{Event}). Consider Fig.~\ref{fig:limitation02}, where the second column provides the follow-up event of the event in the first column. Even if \tabkg{} identifies the correct mapping for the third column to the property \voc{dbo}{locationCity}, we cannot tell if the third column maps to the location of the entity in the first or the second column.
\item Class relations connecting the same classes: We do not have a decision criterion for distinguishing between class relations that connect the same subject and object classes. For example, consider the two object properties \voc{dbo}{leader} and \voc{dbo}{viceLeader} mapped to the second column in Fig.~\ref{fig:limitation01}, both connecting countries to politicians. Only via statistical features extracted from the data table (which may only include the politician's name) it does not appear possible to decide if Kamala Harris is the president or the vice president of the United States.
\end{itemize}

\vspace{-5mm}

\subsubsection{Correlations between Columns and Data Type Relations}

Our data table profiles consist of column profiles, i.e., the features of the single columns are computed in isolation (the same applies to domain profiles and data type relations). Such column profiles can be efficiently computed and added to the dataset profile. However, the dependencies between columns may hold additional knowledge. Consider the running example in Fig. \ref{fig:example_table} where the time in the second column (begin time) does always precede the time in the next column (end time), i.e., there is a correlation between the values in these two columns.

We have decided against the inclusion of correlation features into the domain and data table profiles because of the following reasons: First, correlations are often implicitly captured by the column profiles (e.g., in Fig. \ref{fig:example_table}, the second column's mean value is less than the third column's mean value). Second, the variety of data types requires different correlation and dependency measures that are hard to compare. 
Third, we observed that the number of column pairs heavily exceeds the number of potentially meaningful correlations in our datasets. For example, consider the football dataset where the length of the first names may be compared to the length of last names, team names, the number of goals, \ldots, potentially leading to correlations by chance. 
Fourth, the computation and semantic representation of all possible column combinations are impractical due to the quadratic number of pairwise comparisons.

We also perform data type identification for the columns in isolation. The combination of columns could be used to infer richer data types (e.g., when a date is split into a year, month and day column). We leave such potential optimisations where columns are not just considered in isolation for future work.

\subsubsection{Asymmetry between Domain Profiles and Data Table Profiles}

The domain profile and the data table profile can vary, even though they represent the same knowledge. Consider our running example of weather observations. In the table shown in Fig. \ref{fig:example_table}, three rows refer to the sensor labeled ``S1'' but only two rows refer to the other sensors. When modeled as a knowledge graph following the mapping shown in Fig. \ref{fig:correct_mapping_example}, each sensor is modeled precisely as one node in the knowledge graph. Consequently, the statistical characteristics related to the sensors vary between the data table profile and the domain profile.

\subsubsection{Availability and Representativeness of the Domain Profile}

\tabkg{} relies on the availability of domain-specific data, i.e., a domain knowledge graph. In a recent survey, Abu-Salih provides an overview of approaches to create domain knowledge graphs~\cite{ABUSALIH2021103076}. The profile-based approach of \tabkg{} specifically targets domains of interest which are concise and narrower in scope than existing cross-domain ontologies. As demonstrated in the evaluation, the semantic table interpretation performance decreases when the number of classes and relations increases.

The domain knowledge graph represents available knowledge regarding the domain of interest. We cannot directly measure the representativeness of the knowledge graph. Due to the open-world assumption and, consequently, knowledge graph incompleteness~\cite{arnaout2021negative}, we do not expect the domain knowledge graph to cover the domain knowledge in its entirety. Semantic table interpretation based on a knowledge graph, or its profile, will work for the tables whose content has been adequately represented by the domain knowledge graph, i.e. the data in the domain knowledge graph is representative of the content of these tables. In that case, the domain knowledge and the data table to be interpreted reflect similar value distributions.
If this is not the case, misinterpretation can occur. For example, a data profile solely built from weather observations collected at night is not expected to generalise to daytime observations. In this case, the statistical features of time-related data type relations follow a distribution that is not representative of the whole domain, and thus the profiles will most likely not be matched.

\subsection{Lightweight Semantic Dataset Profiles}

Lightweight semantic profiles generated by \tabkg{} 
can be utilized as a compact domain and dataset representation to complement and enrich existing dataset catalogs. 
Such profiles can be generated automatically from the existing datasets and described using the DCAT\footnote{\url{https://www.w3.org/TR/vocab-dcat-2/}} and the SEAS\footnote{\url{https://ci.mines-stetienne.fr/seas/index.html}} vocabularies to facilitate their reusability. 
We believe that lightweight semantic profiles presented in this article are an essential contribution that can benefit a wide range of semantic applications beyond semantic table interpretation.

Differences in value representations such as value formatting and data types, the representativeness of the domain profile, as well as the dataset characteristics (number of columns in the data table as well as the number of classes and relations in the domain ontology) are the decisive factors to perform semantic table interpretation without instance-lookup and with lightweight profiles.

The profiles of \tabkg{} are extensible. For example, header values, language-specific information or value embeddings can be used as a source of additional profile features in future work.

\section{Related Work}
\label{sec:related_work}

This section provides an overview of related approaches in the areas of dataset profiling and semantic table interpretation.

Given the growth of data available on the Web and in industrial data lakes, there is a high demand for dataset profiling, e.g., for creating data catalogs \cite{neumaier2016automated}. The profile features typically belong to several categories, including statistical observations at the instance and schema level \cite{ben2018rdf, abedjan2015profiling}. Such features are not restricted to the initially defined schemas. For example, Neumaier et al. demonstrate how user interaction and search functionalities profit from the inclusion of spatio-temporal features into a dataset profile \cite{neumaier2019enabling}. For tabular data, other approaches for dataset profile enrichment include the generation of table titles \cite{hancock2019generating} and schema labels \cite{chen2018generating}. The inferred relation-specific rules and observations can further verify the data quality and become part of dataset profiles \cite{schelter2018automating, sejdu2019}.

Recently, approaches to annotate tabular data with concepts from a knowledge base to predict column types gained increased attention. In the following, we introduce approaches for semantic table interpretation and the methods they use.

\textbf{Instance-level lookup.} Most semantic table interpretation tools require access to a target knowledge graph, as they link data table cells to its resources  \cite{ritze2017matching}. Such approaches on the instance-level have recently been driven by the SemTab Challenge \cite{jimenez2020semtab,jimenez2020results}, which explicitly postulates a cell-entity annotation (CEA) task, where labels in data table cells are linked to entities in a target knowledge graph. The subsequent steps of column-type annotation (CTA) and columns-property annotation typically build upon the CEA results.

Several approaches are based on entity lookup (ColNet \cite{chen2019colnet}, MantisTable \cite{cremaschi2020fully}, LinkingPark \cite{chen2020linkingpark}, DAGOBAH \cite{liu2019dagobah,huynh2020dagobah}, MTab \cite{MTab}, T2KMatch \cite{ritze2015matching}, CSV2KG \cite{steenwinckel2019csv2kg}, TableMiner+ \cite{zhang2017effective}, and the work by Zhang et al. \cite{zhang2020novel}), with different (combined) query strategies, including URL matching \cite{steenwinckel2019csv2kg}, (partial) string lookup \cite{cremaschi2020fully,huynh2020dagobah,chen2020linkingpark}, string similarity \cite{MTab,huynh2020dagobah,ritze2015matching,zhang2020novel}, spelling correction \cite{chen2020linkingpark} or the use of named entity linking tools \cite{cremaschi2020fully}. After linking data table cells to resources in the knowledge graphs, the CTA is typically decided through voting or counting \cite{MTab,chen2019colnet,chen2020linkingpark}, ranking \cite{steenwinckel2019csv2kg,cremaschi2020fully,zhang2017effective} or clustering \cite{huynh2020dagobah}. ColNet and TableMiner+ apply learning strategies to reduce the number of lookup tasks required for detecting the class of the entities represented by a column. MantisTable and CSV2KG utilize concept graphs in their ranking to identify the most-specific sub classes.
Also, identifying properties represented by the data tables typically relies on the CEA and additional knowledge graph lookups. For example, MTab does pairwise queries between entities identified in different columns to identify potential entity relations. To identify literal relations, MTab and TableMiner+ row-by-row compute data-type specific similarities between the literals in the target knowledge graph and the cell values. CSV2KG also involves the target ontology in this step. Sherlock \cite{hulsebos2019sherlock} is a system that performs CTA and does not rely on CEA. However, it extracts column features for training a neural network, which is solely trained on DBpedia and explicitly predicts one of the selected DBpedia classes.

The reliance on entity linking with the target knowledge graph makes these approaches unsuitable in settings where data tables only contain previously unseen data, which is a common issue \cite{ritze2016profiling}. Even if the data instances in the data table are not entirely unknown, these approaches do not perform well when the number of matching entities drops \cite{chen2019colnet}. Another thing these approaches have in common is the reliance on a large underlying knowledge base such as DBpedia and stable lookup services. In contrast, \tabkg{} does not require access to the target knowledge graph after the domain profile has been created.

\textbf{Subject column detection.} Several approaches \cite{cremaschi2020fully,zhang2017effective} for semantic table interpretation assume the existence of a subject column, i.e., the main column of the data table where every other column is directly connected to. The subject column detection is typically identified through a set of statistic features. Approaches relying on a subject column do not consider the involvement of any classes which are not directly represented in the data table (for example, consider Fig.~\ref{fig:example_table}, but without the first column). \tabkg{} utilizes a graph-based approach where such class relations can be found.

\textbf{Column titles.} Some data tables come with column titles, which may indicate respective classes or properties. Efthymiou et al. \cite{EfthymiouHRC17} propose a method based on ontology matching, where one column title defines the class label, and other column titles represent property labels. Domain-independent Semantic Labeler \cite{PhamAKS16}, DAGOBAH~\cite{huynh2020dagobah} and TableMiner+~\cite{zhang2017effective} exploit column titles as one of their features. \tabkg{} does not require any column titles. This way, we ensure the generalizability for data tables without headers and language-independence.

\textbf{Data type restrictions.} Data tables contain data of various types, and thus there are approaches specific to some of them. For example, EmbNum+ \cite{NguyenNIT19} transforms data table columns with numeric values into embedding vectors. Alobaid et al. demonstrate that using more fine-grained numeric data types increases semantic table interpretation performance for numeric column values \cite{alobaid2019typology}. For the interpretation of cross-lingual textual values, Luo et al. propose using several translation tools \cite{LuoLCZ18}. \tabkg{} aims at the interpretation of data tables as a whole without restricting to particular data types or languages and thus establishes profiles that do not depend on particular data types or languages.

\textbf{User feedback.} Instead of relying on fully-automated approaches for semantic table interpretation, which may be error-prone due to the challenges involved in this task, manual or semi-automated approaches rely on user feedback. Karma \cite{knoblock2012semi}, Odalic \cite{knap2017towards}, and ASIA \cite{cutrona2019asia} are interactive tools that let users decide on the correctness of suggested table annotations and thus achieve high precision, but demand both time and expertise from the user. \tabkg{} is a fully-automated approach for semantic table interpretation that does not require user interventions.

\textbf{Domain-independent semantic table interpretation.} Domain-independent approaches are not restricted to specific target knowledge graphs. Instead, they learn domain-independent similarity features to generate the mapping. The SemanticTyper \cite{ramnandan2015assigning} scores similarity between columns and data type relations based on handcrafted features for numeric and textual values. Based on similar features, the Domain-independent Semantic Labeler \cite{PhamAKS16} adopts machine learning and handcrafted features to predict the similarity between a column and a class in the domain knowledge graph. Taheriyan et al. \cite{taheriyan2016learning} generate a ranked list of potential column mappings learned from a sample of the domain ontology, which is then presented to the user. While both approaches are flexible concerning the target domain, \tabkg{} aims to use only features present in the dataset profiles.

\section{Conclusion}
\label{sec:conclusion}

This article presented \tabkg{} -- an approach for semantic table interpretation based on lightweight semantic profiles. 
\tabkg{} relies on domain profiles that enrich the relations in a domain ontology and serve as a semantic domain representation. 
\tabkg{} matches these profiles with the tabular data profiles using one-shot learning approach. Our evaluation on five real-world datasets shows that our approach outperforms the baselines for semantic table interpretation. 
In future work, we plan to consider integrating user feedback into the \tabkg{} pipeline to support an extension of the domain ontology in cases where tabular data contains previously unseen relations.

\begin{acks}
 This work is partially funded by the DFG, German Research Foundation (``WorldKG'', 424985896), the Federal Ministry of Education and Research (BMBF), Germany (``Simple-ML'', 01IS18054) and the European Commission (EU H2020, ``smashHit'', 871477).
\end{acks}








\balance
\bibliographystyle{ios1}           
\bibliography{bibliography}        

\begin{thebibliography}{52}
\ifx \bisbn   \undefined \def \bisbn  #1{ISBN #1}\fi
\ifx \binits  \undefined \def \binits#1{#1} \fi
\ifx \bauthor  \undefined \def \bauthor#1{#1} \fi
\ifx \bjtitle  \undefined \def \bjtitle#1{\textit{#1}}\fi
\ifx \batitle  \undefined \def \batitle#1{#1} \fi
\ifx \bctitle  \undefined \def \bctitle#1{#1} \fi
\ifx \bvolume  \undefined \def \bvolume#1{\textbf{#1}}\fi
\ifx \byear  \undefined \def \byear#1{#1} \fi
\ifx \bissue  \undefined \def \bissue#1{#1} \fi
\ifx \bfpage  \undefined \def \bfpage#1{#1} \fi
\ifx \blpage  \undefined \def \blpage #1{#1} \fi
\ifx \burl  \undefined \def \burl#1{#1} \fi
\ifx \doiurl  \undefined \def \doiurl#1{#1} \fi
\ifx \betal  \undefined \def \betal{et al.} \fi
\ifx \binstitute  \undefined \def \binstitute#1{#1} \fi
\ifx \beditor  \undefined \def \beditor#1{#1} \fi
\ifx \bpublisher  \undefined \def \bpublisher#1{#1} \fi
\ifx \bbtitle  \undefined \def \bbtitle#1{\textit{#1}} \fi
\ifx \bedition  \undefined \def \bedition#1{#1} \fi
\ifx \bseriesno  \undefined \def \bseriesno#1{#1} \fi
\ifx \blocation  \undefined \def \blocation#1{#1} \fi
\ifx \bsertitle  \undefined \def \bsertitle#1{#1} \fi
\ifx \bsnm \undefined \def \bsnm#1{#1} \fi
\ifx \bsuffix \undefined \def \bsuffix#1{#1} \fi
\ifx \bparticle \undefined \def \bparticle#1{#1} \fi
\ifx \barticle \undefined \def \barticle#1{#1} \fi
\ifx \botherref \undefined \def \botherref #1{#1} \fi
\ifx \url \undefined \def \url#1{#1} \fi
\ifx \bchapter \undefined \def \bchapter#1{#1} \fi
\ifx \bbook \undefined \def \bbook#1{#1} \fi
\ifx \bcomment \undefined \def \bcomment#1{#1} \fi
\ifx \oauthor \undefined \def \oauthor#1{#1} \fi
\ifx \citeauthoryear \undefined \def \citeauthoryear#1{#1} \fi
\ifx \texttildelow  \undefined \def \texttildelow{\symbol{126}} \fi
\def \endbibitem {}
\ifx \bconflocation  \undefined \def \bconflocation#1{#1} \fi

\bibitem{cafarella2018ten}
\begin{barticle}
\bauthor{\binits{M.J.}~\bsnm{Cafarella}},
\bauthor{\binits{A.Y.}~\bsnm{Halevy}},
\bauthor{\binits{H.}~\bsnm{Lee}},
\bauthor{\binits{J.}~\bsnm{Madhavan}},
\bauthor{\binits{C.}~\bsnm{Yu}},
\bauthor{\binits{D.Z.}~\bsnm{Wang}} and
\bauthor{\binits{E.}~\bsnm{Wu}},
\batitle{{Ten Years of WebTables}},
\bjtitle{Proc. {VLDB} Endow.}
\bvolume{11}(\bissue{12})
(\byear{2018}),
\bfpage{2140}--\blpage{2149},
\bcomment{DOI: \url{https://doi.org/10.14778/3229863.3240492}}.
\end{barticle}
\endbibitem

\bibitem{ritze2017matching}
\begin{bchapter}
\bauthor{\binits{D.}~\bsnm{Ritze}} and
\bauthor{\binits{C.}~\bsnm{Bizer}},
\bctitle{{Matching Web Tables to DBpedia - a Feature Utility Study}},
in: \bbtitle{Proceedings of the 20th International Conference on Extending
  Database Technology, {EDBT} 2017, Venice, Italy, March 21-24, 2017},
\bpublisher{OpenProceedings.org},
\byear{2017},
pp.~\bfpage{210}--\blpage{221},
\bcomment{DOI: \url{https://doi.org/10.5441/002/edbt.2017.20}}.
\end{bchapter}
\endbibitem

\bibitem{mitlohner2016characteristics}
\begin{bchapter}
\bauthor{\binits{J.}~\bsnm{Mitl{\"{o}}hner}},
\bauthor{\binits{S.}~\bsnm{Neumaier}},
\bauthor{\binits{J.}~\bsnm{Umbrich}} and
\bauthor{\binits{A.}~\bsnm{Polleres}},
\bctitle{{Characteristics of Open Data CSV Files}},
in: \bbtitle{Proceedings of the 2nd International Conference on Open and Big
  Data, {OBD} 2016, Vienna, Austria, August 22-24, 2016},
\bpublisher{{IEEE} Computer Society},
\byear{2016},
pp.~\bfpage{72}--\blpage{79},
\bcomment{DOI: \url{https://doi.org/10.1109/OBD.2016.18}}.
\end{bchapter}
\endbibitem

\bibitem{gottschalk2019simple}
\begin{bchapter}
\bauthor{\binits{S.}~\bsnm{Gottschalk}},
\bauthor{\binits{N.}~\bsnm{Tempelmeier}},
\bauthor{\binits{G.}~\bsnm{Kniesel}},
\bauthor{\binits{V.}~\bsnm{Iosifidis}},
\bauthor{\binits{B.}~\bsnm{Fetahu}} and
\bauthor{\binits{E.}~\bsnm{Demidova}},
\bctitle{{Simple-ML: Towards a Framework for Semantic Data Analytics
  Workflows}},
in: \bbtitle{Semantic Systems. The Power of {AI} and Knowledge Graphs - 15th
  International Conference, SEMANTiCS 2019, Karlsruhe, Germany, September 9-12,
  2019, Proceedings},
\bsertitle{Lecture Notes in Computer Science},
Vol.~\bseriesno{11702},
\bpublisher{Springer},
\byear{2019},
pp.~\bfpage{359}--\blpage{366},
\bcomment{DOI: \url{https://doi.org/10.1007/978-3-030-33220-4\_26}}.
\end{bchapter}
\endbibitem

\bibitem{garda2019semantics}
\begin{bchapter}
\bauthor{\binits{M.}~\bsnm{Garda}},
\bctitle{{A Semantics-Enabled Approach for Data Lake Exploration Services}},
in: \bbtitle{2019 {IEEE} World Congress on Services, {SERVICES} 2019, Milan,
  Italy, July 8-13, 2019},
\bpublisher{{IEEE}},
\byear{2019},
pp.~\bfpage{327}--\blpage{330},
\bcomment{DOI: \url{https://doi.org/10.1109/SERVICES.2019.00091}}.
\end{bchapter}
\endbibitem

\bibitem{pomp2018applying}
\begin{barticle}
\bauthor{\binits{A.}~\bsnm{Pomp}},
\bauthor{\binits{A.}~\bsnm{Paulus}},
\bauthor{\binits{A.}~\bsnm{Kirmse}},
\bauthor{\binits{V.}~\bsnm{Kraus}} and
\bauthor{\binits{T.}~\bsnm{Meisen}},
\batitle{{Applying Semantics to Reduce the Time to Analytics within Complex
  Heterogeneous Infrastructures}},
\bjtitle{Technologies}
\bvolume{6}(\bissue{3})
(\byear{2018}),
\bfpage{86},
\bcomment{DOI: \url{https://doi.org/10.3390/technologies6030086}}.
\end{barticle}
\endbibitem

\bibitem{hartenfels2017type}
\begin{bchapter}
\bauthor{\binits{C.}~\bsnm{Hartenfels}},
\bauthor{\binits{M.}~\bsnm{Leinberger}},
\bauthor{\binits{R.}~\bsnm{L{\"{a}}mmel}} and
\bauthor{\binits{S.}~\bsnm{Staab}},
\bctitle{{Type-Safe Programming with OWL in Semantics4J.}},
in: \bbtitle{Proceedings of the {ISWC} 2017 Posters {\&} Demonstrations and
  Industry Tracks co-located with 16th International Semantic Web Conference
  {(ISWC} 2017), Vienna, Austria, October 23rd - to - 25th, 2017},
\bsertitle{{CEUR} Workshop Proceedings},
Vol.~\bseriesno{1963},
\bpublisher{CEUR-WS.org},
\byear{2017}.
\url{http://ceur-ws.org/Vol-1963/paper549.pdf}.
\end{bchapter}
\endbibitem

\bibitem{lecue2019role}
\begin{barticle}
\bauthor{\binits{F.}~\bsnm{L{\'{e}}cu{\'{e}}}},
\batitle{{On the Role of Knowledge Graphs in Explainable AI}},
\bjtitle{Semantic Web}
\bvolume{11}(\bissue{1})
(\byear{2020}),
\bfpage{41}--\blpage{51},
\bcomment{DOI: \url{https://doi.org/10.3233/SW-190374}}.
\url{https://doi.org/10.3233/SW-190374}.
\end{barticle}
\endbibitem

\bibitem{lehmann2017distributed}
\begin{bchapter}
\bauthor{\binits{J.}~\bsnm{Lehmann}},
\bauthor{\binits{G.}~\bsnm{Sejdiu}},
\bauthor{\binits{L.}~\bsnm{B{\"{u}}hmann}},
\bauthor{\binits{P.}~\bsnm{Westphal}},
\bauthor{\binits{C.}~\bsnm{Stadler}},
\bauthor{\binits{I.}~\bsnm{Ermilov}},
\bauthor{\binits{S.}~\bsnm{Bin}},
\bauthor{\binits{N.}~\bsnm{Chakraborty}},
\bauthor{\binits{M.}~\bsnm{Saleem}},
\bauthor{\binits{A.N.}~\bsnm{Ngomo}} and
\bauthor{\binits{H.}~\bsnm{Jabeen}},
\bctitle{{Distributed Semantic Analytics using the SANSA Stack}},
in: \bbtitle{The Semantic Web - {ISWC} 2017 - 16th International Semantic Web
  Conference, Vienna, Austria, October 21-25, 2017, Proceedings, Part {II}},
\bsertitle{Lecture Notes in Computer Science},
Vol.~\bseriesno{10588},
\bpublisher{Springer},
\byear{2017},
pp.~\bfpage{147}--\blpage{155},
\bcomment{DOI: \url{https://doi.org/10.1007/978-3-319-68204-4\_15}}.
\end{bchapter}
\endbibitem

\bibitem{MTab}
\begin{bchapter}
\bauthor{\binits{P.}~\bsnm{Nguyen}},
\bauthor{\binits{N.}~\bsnm{Kertkeidkachorn}},
\bauthor{\binits{R.}~\bsnm{Ichise}} and
\bauthor{\binits{H.}~\bsnm{Takeda}},
\bctitle{MTab: Matching Tabular Data to Knowledge Graph using Probability
  Models},
in: \bbtitle{Proceedings of the Semantic Web Challenge on Tabular Data to
  Knowledge Graph Matching co-located with the 18th International Semantic Web
  Conference, SemTab@ISWC 2019, Auckland, New Zealand, October 30, 2019},
\bsertitle{{CEUR} Workshop Proceedings},
Vol.~\bseriesno{2553},
\bpublisher{CEUR-WS.org},
\byear{2019},
pp.~\bfpage{7}--\blpage{14}.
\url{http://ceur-ws.org/Vol-2553/paper2.pdf}.
\end{bchapter}
\endbibitem

\bibitem{cremaschi2020fully}
\begin{barticle}
\bauthor{\binits{M.}~\bsnm{Cremaschi}},
\bauthor{\binits{F.D.}~\bsnm{Paoli}},
\bauthor{\binits{A.}~\bsnm{Rula}} and
\bauthor{\binits{B.}~\bsnm{Spahiu}},
\batitle{{A Fully Automated Approach to a Complete Semantic Table
  Interpretation}},
\bjtitle{Future Gener. Comput. Syst.}
\bvolume{112}
(\byear{2020}),
\bfpage{478}--\blpage{500},
\bcomment{DOI: \url{https://doi.org/10.1016/j.future.2020.05.019}}.
\end{barticle}
\endbibitem

\bibitem{chen2019colnet}
\begin{bchapter}
\bauthor{\binits{J.}~\bsnm{Chen}},
\bauthor{\binits{E.}~\bsnm{Jim{\'{e}}nez{-}Ruiz}},
\bauthor{\binits{I.}~\bsnm{Horrocks}} and
\bauthor{\binits{C.}~\bsnm{Sutton}},
\bctitle{{ColNet: Embedding the Semantics of Web Tables for Column Type
  Prediction}},
in: \bbtitle{Proceedings of the Thirty-Third Conference on Artificial
  Intelligence, {AAAI} 2019},
\bpublisher{{AAAI} Press},
\byear{2019},
pp.~\bfpage{29}--\blpage{36},
\bcomment{DOI: \url{https://doi.org/10.1609/aaai.v33i01.330129}}.
\end{bchapter}
\endbibitem

\bibitem{auer2007dbpedia}
\begin{bchapter}
\bauthor{\binits{S.}~\bsnm{Auer}},
\bauthor{\binits{C.}~\bsnm{Bizer}},
\bauthor{\binits{G.}~\bsnm{Kobilarov}},
\bauthor{\binits{J.}~\bsnm{Lehmann}},
\bauthor{\binits{R.}~\bsnm{Cyganiak}} and
\bauthor{\binits{Z.G.}~\bsnm{Ives}},
\bctitle{{DBpedia: A Nucleus for a Web of Open Data}},
in: \bbtitle{The Semantic Web, 6th International Semantic Web Conference, 2nd
  Asian Semantic Web Conference, {ISWC} 2007 + {ASWC} 2007, Busan, Korea,
  November 11-15, 2007},
\bsertitle{Lecture Notes in Computer Science},
Vol.~\bseriesno{4825},
\bpublisher{Springer},
\byear{2007},
pp.~\bfpage{722}--\blpage{735},
\bcomment{DOI: \url{https://doi.org/10.1007/978-3-540-76298-0\_52}}.
\end{bchapter}
\endbibitem

\bibitem{ritze2016profiling}
\begin{bchapter}
\bauthor{\binits{D.}~\bsnm{Ritze}},
\bauthor{\binits{O.}~\bsnm{Lehmberg}},
\bauthor{\binits{Y.}~\bsnm{Oulabi}} and
\bauthor{\binits{C.}~\bsnm{Bizer}},
\bctitle{Profiling the Potential of Web Tables for Augmenting Cross-domain
  Knowledge Bases},
in: \bbtitle{Proceedings of the 25th International Conference on World Wide
  Web, {WWW} 2016, Montreal, Canada, April 11 - 15, 2016},
\bpublisher{{ACM}},
\byear{2016},
pp.~\bfpage{251}--\blpage{261},
\bcomment{DOI: \url{https://doi.org/10.1145/2872427.2883017}}.
\end{bchapter}
\endbibitem

\bibitem{ben2018rdf}
\begin{barticle}
\bauthor{\binits{M.B.}~\bsnm{Ellefi}},
\bauthor{\binits{Z.}~\bsnm{Bellahsene}},
\bauthor{\binits{J.G.}~\bsnm{Breslin}},
\bauthor{\binits{E.}~\bsnm{Demidova}},
\bauthor{\binits{S.}~\bsnm{Dietze}},
\bauthor{\binits{J.}~\bsnm{Szymanski}} and
\bauthor{\binits{K.}~\bsnm{Todorov}},
\batitle{{RDF Dataset Profiling--a Survey of Features, Methods, Vocabularies
  and Applications}},
\bjtitle{Semantic Web}
\bvolume{9}(\bissue{5})
(\byear{2018}),
\bfpage{677}--\blpage{705},
\bcomment{DOI: \url{https://doi.org/10.3233/SW-180294}}.
\end{barticle}
\endbibitem

\bibitem{alobaid2019typology}
\begin{barticle}
\bauthor{\binits{A.}~\bsnm{Alobaid}},
\bauthor{\binits{E.}~\bsnm{Kacprzak}} and
\bauthor{\binits{{\'{O}}.}~\bsnm{Corcho}},
\batitle{{Typology-based Semantic Labeling of Numeric Tabular Data}},
\bjtitle{Semantic Web}
\bvolume{12}(\bissue{1})
(\byear{2021}),
\bfpage{5}--\blpage{20},
\bcomment{DOI: \url{https://doi.org/10.3233/SW-200397}}.
\end{barticle}
\endbibitem

\bibitem{harth2010data}
\begin{bchapter}
\bauthor{\binits{A.}~\bsnm{Harth}},
\bauthor{\binits{K.}~\bsnm{Hose}},
\bauthor{\binits{M.}~\bsnm{Karnstedt}},
\bauthor{\binits{A.}~\bsnm{Polleres}},
\bauthor{\binits{K.}~\bsnm{Sattler}} and
\bauthor{\binits{J.}~\bsnm{Umbrich}},
\bctitle{{Data Summaries for On-demand Queries over Linked Data}},
in: \bbtitle{Proceedings of the 19th International Conference on World Wide
  Web, {WWW} 2010, Raleigh, North Carolina, USA, April 26-30, 2010},
\bpublisher{{ACM}},
\byear{2010},
pp.~\bfpage{411}--\blpage{420},
\bcomment{DOI: \url{https://doi.org/10.1145/1772690.1772733}}.
\end{bchapter}
\endbibitem

\bibitem{koch2015siamese}
\begin{bchapter}
\bauthor{\binits{G.}~\bsnm{Koch}},
\bauthor{\binits{R.}~\bsnm{Zemel}} and
\bauthor{\binits{R.}~\bsnm{Salakhutdinov}},
\bctitle{{Siamese Neural Networks for One-shot Image Recognition}},
in: \bbtitle{Proceedings of the Deep Learning Workshop, International
  Conference on Machine Learning '15},
Vol.~\bseriesno{2},
\byear{2015}.
\end{bchapter}
\endbibitem

\bibitem{hulsebos2019sherlock}
\begin{bchapter}
\bauthor{\binits{M.}~\bsnm{Hulsebos}},
\bauthor{\binits{K.Z.}~\bsnm{Hu}},
\bauthor{\binits{M.A.}~\bsnm{Bakker}},
\bauthor{\binits{E.}~\bsnm{Zgraggen}},
\bauthor{\binits{A.}~\bsnm{Satyanarayan}},
\bauthor{\binits{T.}~\bsnm{Kraska}},
\bauthor{\binits{{\c{C}}.}~\bsnm{Demiralp}} and
\bauthor{\binits{C.A.}~\bsnm{Hidalgo}},
\bctitle{{Sherlock: A Deep Learning Approach to Semantic Data Type Detection}},
in: \bbtitle{Proceedings of the 25th {ACM} {SIGKDD} International Conference on
  Knowledge Discovery {\&} Data Mining, {KDD} 2019, Anchorage, AK, USA, August
  4-8, 2019},
\bpublisher{{ACM}},
\byear{2019},
pp.~\bfpage{1500}--\blpage{1508},
\bcomment{DOI: \url{https://doi.org/10.1145/3292500.3330993}}.
\end{bchapter}
\endbibitem

\bibitem{PhamAKS16}
\begin{bchapter}
\bauthor{\binits{M.}~\bsnm{Pham}},
\bauthor{\binits{S.}~\bsnm{Alse}},
\bauthor{\binits{C.A.}~\bsnm{Knoblock}} and
\bauthor{\binits{P.A.}~\bsnm{Szekely}},
\bctitle{Semantic Labeling: {A} Domain-Independent Approach},
in: \bbtitle{The Semantic Web - {ISWC} 2016 - 15th International Semantic Web
  Conference, Kobe, Japan, October 17-21, 2016, Proceedings, Part {I}},
\bsertitle{Lecture Notes in Computer Science},
Vol.~\bseriesno{9981},
\byear{2016},
pp.~\bfpage{446}--\blpage{462},
\bcomment{DOI: \url{https://doi.org/10.1007/978-3-319-46523-4\_27}}.
\end{bchapter}
\endbibitem

\bibitem{Dimou:2014}
\begin{bchapter}
\bauthor{\binits{A.}~\bsnm{Dimou}},
\bauthor{\binits{M.V.}~\bsnm{Sande}},
\bauthor{\binits{P.}~\bsnm{Colpaert}},
\bauthor{\binits{R.}~\bsnm{Verborgh}},
\bauthor{\binits{E.}~\bsnm{Mannens}} and
\bauthor{\binits{R.V.}~\bsnm{de~Walle}},
\bctitle{{RML: A Generic Language for Integrated RDF Mappings of Heterogeneous
  Data}},
in: \bbtitle{Proceedings of the Workshop on Linked Data on the Web co-located
  with the 23rd International World Wide Web Conference {(WWW} 2014), Seoul,
  Korea, April 8, 2014},
\bsertitle{{CEUR} Workshop Proceedings},
Vol.~\bseriesno{1184},
\bpublisher{CEUR-WS.org},
\byear{2014}.
\url{http://ceur-ws.org/Vol-1184/ldow2014\_paper\_01.pdf}.
\end{bchapter}
\endbibitem

\bibitem{wkbwkt}
\begin{botherref}
{Information Technology – Database Languages – SQL Multimedia and
  Application Packages – Part 3: Spatial},
Standard,
International Organization for Standardization,
2016.
\url{https://www.iso.org/standard/60343.html}.
\end{botherref}
\endbibitem

\bibitem{heise2013scalable}
\begin{barticle}
\bauthor{\binits{A.}~\bsnm{Heise}},
\bauthor{\binits{J.}~\bsnm{Quian{\'{e}}{-}Ruiz}},
\bauthor{\binits{Z.}~\bsnm{Abedjan}},
\bauthor{\binits{A.}~\bsnm{Jentzsch}} and
\bauthor{\binits{F.}~\bsnm{Naumann}},
\batitle{Scalable Discovery of Unique Column Combinations},
\bjtitle{Proc. {VLDB} Endow.}
\bvolume{7}(\bissue{4})
(\byear{2013}),
\bfpage{301}--\blpage{312},
\bcomment{DOI: \url{https://doi.org/10.14778/2732240.2732248}}.
\url{http://www.vldb.org/pvldb/vol7/p301-heise.pdf}.
\end{barticle}
\endbibitem

\bibitem{jimenez2020semtab}
\begin{botherref}
\oauthor{\binits{E.}~\bsnm{Jim{\'{e}}nez{-}Ruiz}},
\oauthor{\binits{O.}~\bsnm{Hassanzadeh}},
\oauthor{\binits{V.}~\bsnm{Efthymiou}},
\oauthor{\binits{J.}~\bsnm{Chen}} and
\oauthor{\binits{K.}~\bsnm{Srinivas}},
SemTab 2019: Resources to Benchmark Tabular Data to Knowledge Graph Matching
  Systems
\textbf{12123}
(2020),
514--530,
DOI: \url{https://doi.org/10.1007/978-3-030-49461-2\_30}.
\url{https://doi.org/10.1007/978-3-030-49461-2\_30}.
\end{botherref}
\endbibitem

\bibitem{taheriyan2016leveraging}
\begin{bchapter}
\bauthor{\binits{M.}~\bsnm{Taheriyan}},
\bauthor{\binits{C.A.}~\bsnm{Knoblock}},
\bauthor{\binits{P.A.}~\bsnm{Szekely}} and
\bauthor{\binits{J.L.}~\bsnm{Ambite}},
\bctitle{Leveraging Linked Data to Discover Semantic Relations Within Data
  Sources},
in: \bbtitle{The Semantic Web - {ISWC} 2016 - 15th International Semantic Web
  Conference, Kobe, Japan, October 17-21, 2016, Proceedings, Part {I}},
\bsertitle{Lecture Notes in Computer Science},
Vol.~\bseriesno{9981},
\byear{2016},
pp.~\bfpage{549}--\blpage{565},
\bcomment{DOI: \url{10.1007/978-3-319-46523-4\_33}}.
\url{https://doi.org/10.1007/978-3-319-46523-4\_33}.
\end{bchapter}
\endbibitem

\bibitem{ritze2015matching}
\begin{bchapter}
\bauthor{\binits{D.}~\bsnm{Ritze}},
\bauthor{\binits{O.}~\bsnm{Lehmberg}} and
\bauthor{\binits{C.}~\bsnm{Bizer}},
\bctitle{Matching {HTML} Tables to DBpedia},
in: \bbtitle{Proceedings of the 5th International Conference on Web
  Intelligence, Mining and Semantics, {WIMS} 2015, Larnaca, Cyprus, July 13-15,
  2015},
\bpublisher{{ACM}},
\byear{2015},
pp.~\bfpage{10:1}--\blpage{10:6},
\bcomment{DOI: \url{https://doi.org/10.1145/2797115.2797118}}.
\end{bchapter}
\endbibitem

\bibitem{chen2020linkingpark}
\begin{botherref}
\oauthor{\binits{S.}~\bsnm{Chen}},
\oauthor{\binits{A.}~\bsnm{Karaoglu}},
\oauthor{\binits{C.}~\bsnm{Negreanu}},
\oauthor{\binits{T.}~\bsnm{Ma}},
\oauthor{\binits{J.}~\bsnm{Yao}},
\oauthor{\binits{J.}~\bsnm{Williams}},
\oauthor{\binits{A.}~\bsnm{Gordon}} and
\oauthor{\binits{C.}~\bsnm{Lin}},
LinkingPark: An Integrated Approach for Semantic Table Interpretation
\textbf{2775}
(2020),
65--74.
\url{http://ceur-ws.org/Vol-2775/paper7.pdf}.
\end{botherref}
\endbibitem

\bibitem{zhang2017effective}
\begin{barticle}
\bauthor{\binits{Z.}~\bsnm{Zhang}},
\batitle{Effective and efficient Semantic Table Interpretation using
  TableMiner\({}^{\mbox{+}}\)},
\bjtitle{Semantic Web}
\bvolume{8}(\bissue{6})
(\byear{2017}),
\bfpage{921}--\blpage{957},
\bcomment{DOI: \url{https://doi.org/10.3233/SW-160242}}.
\end{barticle}
\endbibitem

\bibitem{zhang2020web}
\begin{barticle}
\bauthor{\binits{S.}~\bsnm{Zhang}} and
\bauthor{\binits{K.}~\bsnm{Balog}},
\batitle{{Web Table Extraction, Retrieval, and Augmentation: A Survey}},
\bjtitle{ACM Transactions on Intelligent Systems and Technology (TIST)}
\bvolume{11}(\bissue{2})
(\byear{2020}),
\bfpage{1}--\blpage{35},
\bcomment{DOI: \url{https://doi.org/10.1145/3372117}}.
\end{barticle}
\endbibitem

\bibitem{ramnandan2015assigning}
\begin{bchapter}
\bauthor{\binits{S.K.}~\bsnm{Ramnandan}},
\bauthor{\binits{A.}~\bsnm{Mittal}},
\bauthor{\binits{C.A.}~\bsnm{Knoblock}} and
\bauthor{\binits{P.A.}~\bsnm{Szekely}},
\bctitle{Assigning Semantic Labels to Data Sources},
in: \bbtitle{The Semantic Web. Latest Advances and New Domains - 12th European
  Semantic Web Conference, {ESWC} 2015, Portoroz, Slovenia, May 31 - June 4,
  2015. Proceedings},
\bsertitle{Lecture Notes in Computer Science},
Vol.~\bseriesno{9088},
\bpublisher{Springer},
\byear{2015},
pp.~\bfpage{403}--\blpage{417},
\bcomment{DOI: \url{10.1007/978-3-319-18818-8\_25}}.
\end{bchapter}
\endbibitem

\bibitem{zhang2020novel}
\begin{bchapter}
\bauthor{\binits{S.}~\bsnm{Zhang}},
\bauthor{\binits{E.}~\bsnm{Meij}},
\bauthor{\binits{K.}~\bsnm{Balog}} and
\bauthor{\binits{R.}~\bsnm{Reinanda}},
\bctitle{Novel Entity Discovery from Web Tables},
in: \bbtitle{Proceedings of the Web Conference 2020 (TheWebConf), Taipei,
  Taiwan, April 20-24, 2020},
\bpublisher{{ACM} / {IW3C2}},
\byear{2020},
pp.~\bfpage{1298}--\blpage{1308},
\bcomment{DOI: \url{https://doi.org/10.1145/3366423.3380205}}.
\end{bchapter}
\endbibitem

\bibitem{EfthymiouHRC17}
\begin{bchapter}
\bauthor{\binits{V.}~\bsnm{Efthymiou}},
\bauthor{\binits{O.}~\bsnm{Hassanzadeh}},
\bauthor{\binits{M.}~\bsnm{Rodriguez{-}Muro}} and
\bauthor{\binits{V.}~\bsnm{Christophides}},
\bctitle{{Matching Web Tables with Knowledge Base Entities: From Entity Lookups
  to Entity Embeddings}},
in: \bbtitle{The Semantic Web - {ISWC} 2017 - 16th International Semantic Web
  Conference, Vienna, Austria, October 21-25, 2017, Proceedings, Part {I}},
\bsertitle{Lecture Notes in Computer Science},
Vol.~\bseriesno{10587},
\bpublisher{Springer},
\byear{2017},
pp.~\bfpage{260}--\blpage{277},
\bcomment{DOI: \url{10.1007/978-3-319-68288-4\_16}}.
\end{bchapter}
\endbibitem

\bibitem{hsiao2019malware}
\begin{botherref}
\oauthor{\binits{S.}~\bsnm{Hsiao}},
\oauthor{\binits{D.}~\bsnm{Kao}},
\oauthor{\binits{Z.}~\bsnm{Liu}} and
\oauthor{\binits{R.}~\bsnm{Tso}},
{Malware Image Classification using One-shot Learning with Siamese Networks}
\textbf{159}
(2019),
1863--1871,
DOI: \url{https://doi.org/10.1016/j.procs.2019.09.358}.
\end{botherref}
\endbibitem

\bibitem{ABUSALIH2021103076}
\begin{barticle}
\bauthor{\binits{B.}~\bsnm{Abu-Salih}},
\batitle{{Domain-specific Knowledge Graphs: A Survey}},
\bjtitle{Journal of Network and Computer Applications}
\bvolume{185}
(\byear{2021}),
\bfpage{103076},
\bcomment{DOI: \url{https://doi.org/10.1016/j.jnca.2021.103076}}.
\url{https://www.sciencedirect.com/science/article/pii/S1084804521000990}.
\end{barticle}
\endbibitem

\bibitem{arnaout2021negative}
\begin{bchapter}
\bauthor{\binits{H.}~\bsnm{Arnaout}},
\bauthor{\binits{S.}~\bsnm{Razniewski}},
\bauthor{\binits{G.}~\bsnm{Weikum}} and
\bauthor{\binits{J.Z.}~\bsnm{Pan}},
\bctitle{{Negative Knowledge for Open-world Wikidata}},
in: \bbtitle{Companion of The Web Conference 2021, Virtual Event / Ljubljana,
  Slovenia, April 19-23, 2021},
\byear{2021},
pp.~\bfpage{544}--\blpage{551},
\bcomment{DOI: \url{https://doi.org/10.1145/3442442.3452339}}.
\end{bchapter}
\endbibitem

\bibitem{neumaier2016automated}
\begin{barticle}
\bauthor{\binits{S.}~\bsnm{Neumaier}},
\bauthor{\binits{J.}~\bsnm{Umbrich}} and
\bauthor{\binits{A.}~\bsnm{Polleres}},
\batitle{{Automated Quality Assessment of Metadata across Open Data Portals}},
\bjtitle{{ACM} J. Data Inf. Qual.}
\bvolume{8}(\bissue{1})
(\byear{2016}),
\bfpage{2:1}--\blpage{2:29},
\bcomment{DOI: \url{https://doi.org/10.1145/2964909}}.
\url{https://doi.org/10.1145/2964909}.
\end{barticle}
\endbibitem

\bibitem{abedjan2015profiling}
\begin{barticle}
\bauthor{\binits{Z.}~\bsnm{Abedjan}},
\bauthor{\binits{L.}~\bsnm{Golab}} and
\bauthor{\binits{F.}~\bsnm{Naumann}},
\batitle{{Profiling Relational Data: a Survey}},
\bjtitle{{VLDB} J.}
\bvolume{24}(\bissue{4})
(\byear{2015}),
\bfpage{557}--\blpage{581},
\bcomment{DOI: \url{https://doi.org/10.1007/s00778-015-0389-y}}.
\end{barticle}
\endbibitem

\bibitem{neumaier2019enabling}
\begin{barticle}
\bauthor{\binits{S.}~\bsnm{Neumaier}} and
\bauthor{\binits{A.}~\bsnm{Polleres}},
\batitle{Enabling Spatio-Temporal Search in Open Data},
\bjtitle{J. Web Semant.}
\bvolume{55}
(\byear{2019}),
\bfpage{21}--\blpage{36},
\bcomment{DOI: \url{https://doi.org/10.1016/j.websem.2018.12.007}}.
\end{barticle}
\endbibitem

\bibitem{hancock2019generating}
\begin{bchapter}
\bauthor{\binits{B.}~\bsnm{Hancock}},
\bauthor{\binits{H.}~\bsnm{Lee}} and
\bauthor{\binits{C.}~\bsnm{Yu}},
\bctitle{Generating Titles for Web Tables},
in: \bbtitle{Proceedings of the World Wide Web Conference, {TheWebConf} 2019,
  San Francisco, CA, USA, May 13-17, 2019},
\bpublisher{{ACM}},
\byear{2019},
pp.~\bfpage{638}--\blpage{647},
\bcomment{DOI: \url{https://doi.org/10.1145/3308558.3313399}}.
\end{bchapter}
\endbibitem

\bibitem{chen2018generating}
\begin{bchapter}
\bauthor{\binits{Z.}~\bsnm{Chen}},
\bauthor{\binits{H.}~\bsnm{Jia}},
\bauthor{\binits{J.}~\bsnm{Heflin}} and
\bauthor{\binits{B.D.}~\bsnm{Davison}},
\bctitle{{Generating Schema Labels through Dataset Content Analysis}},
in: \bbtitle{Companion of the The Web Conference 2018 on The Web Conference
  2018, {WWW} 2018, Lyon , France, April 23-27, 2018},
\bpublisher{{ACM}},
\byear{2018},
pp.~\bfpage{1515}--\blpage{1522},
\bcomment{DOI: \url{https://doi.org/10.1145/3184558.3191601}}.
\end{bchapter}
\endbibitem

\bibitem{schelter2018automating}
\begin{barticle}
\bauthor{\binits{S.}~\bsnm{Schelter}},
\bauthor{\binits{D.}~\bsnm{Lange}},
\bauthor{\binits{P.}~\bsnm{Schmidt}},
\bauthor{\binits{M.}~\bsnm{Celikel}},
\bauthor{\binits{F.}~\bsnm{Bie{\ss}mann}} and
\bauthor{\binits{A.}~\bsnm{Grafberger}},
\batitle{{Automating Large-scale Data Quality Verification}},
\bjtitle{Proc. {VLDB} Endow.}
\bvolume{11}(\bissue{12})
(\byear{2018}),
\bfpage{1781}--\blpage{1794},
\bcomment{DOI: \url{https://doi.org/10.14778/3229863.3229867}}.
\end{barticle}
\endbibitem

\bibitem{sejdu2019}
\begin{bchapter}
\bauthor{\binits{G.}~\bsnm{Sejdiu}},
\bauthor{\binits{A.}~\bsnm{Rula}},
\bauthor{\binits{J.}~\bsnm{Lehmann}} and
\bauthor{\binits{H.}~\bsnm{Jabeen}},
\bctitle{{A Scalable Framework for Quality Assessment of RDF Datasets}},
in: \bbtitle{The Semantic Web - {ISWC} 2019 - 18th International Semantic Web
  Conference, Auckland, New Zealand, October 26-30, 2019, Proceedings, Part
  {II}},
\bsertitle{Lecture Notes in Computer Science},
Vol.~\bseriesno{11779},
\bpublisher{Springer},
\byear{2019},
pp.~\bfpage{261}--\blpage{276},
\bcomment{DOI: \url{https://doi.org/10.1007/978-3-030-30796-7\_17}}.
\end{bchapter}
\endbibitem

\bibitem{jimenez2020results}
\begin{bchapter}
\bauthor{\binits{E.}~\bsnm{Jim{\'{e}}nez{-}Ruiz}},
\bauthor{\binits{O.}~\bsnm{Hassanzadeh}},
\bauthor{\binits{V.}~\bsnm{Efthymiou}},
\bauthor{\binits{J.}~\bsnm{Chen}},
\bauthor{\binits{K.}~\bsnm{Srinivas}} and
\bauthor{\binits{V.}~\bsnm{Cutrona}},
\bctitle{Results of SemTab 2020},
in: \bbtitle{Proceedings of the Semantic Web Challenge on Tabular Data to
  Knowledge Graph Matching (SemTab 2020) co-located with the 19th International
  Semantic Web Conference {(ISWC} 2020), Virtual conference (originally planned
  to be in Athens, Greece), November 5, 2020},
\bsertitle{{CEUR} Workshop Proceedings},
Vol.~\bseriesno{2775},
\bpublisher{CEUR-WS.org},
\byear{2020},
pp.~\bfpage{1}--\blpage{8}.
\url{http://ceur-ws.org/Vol-2775/paper0.pdf}.
\end{bchapter}
\endbibitem

\bibitem{liu2019dagobah}
\begin{botherref}
\oauthor{\binits{Y.}~\bsnm{Chabot}},
\oauthor{\binits{T.}~\bsnm{Labb{\'{e}}}},
\oauthor{\binits{J.}~\bsnm{Liu}} and
\oauthor{\binits{R.}~\bsnm{Troncy}},
{DAGOBAH: An End-to-End Context-Free Tabular Data Semantic Annotation System}
\textbf{2553}
(2019),
41--48.
\url{http://ceur-ws.org/Vol-2553/paper6.pdf}.
\end{botherref}
\endbibitem

\bibitem{huynh2020dagobah}
\begin{botherref}
\oauthor{\binits{V.}~\bsnm{Huynh}},
\oauthor{\binits{J.}~\bsnm{Liu}},
\oauthor{\binits{Y.}~\bsnm{Chabot}},
\oauthor{\binits{T.}~\bsnm{Labb{\'{e}}}},
\oauthor{\binits{P.}~\bsnm{Monnin}} and
\oauthor{\binits{R.}~\bsnm{Troncy}},
{DAGOBAH: Enhanced Scoring Algorithms for Scalable Annotations of Tabular Data}
\textbf{2775}
(2020),
27--39.
\url{http://ceur-ws.org/Vol-2775/paper3.pdf}.
\end{botherref}
\endbibitem

\bibitem{steenwinckel2019csv2kg}
\begin{bchapter}
\bauthor{\binits{B.}~\bsnm{Steenwinckel}},
\bauthor{\binits{G.}~\bsnm{Vandewiele}},
\bauthor{\binits{F.}~\bsnm{De~Turck}} and
\bauthor{\binits{F.}~\bsnm{Ongenae}},
\bctitle{{CSV2KG: Transforming Tabular Data into Semantic Knowledge}},
in: \bbtitle{Proceedings of the Semantic Web Challenge on Tabular Data to
  Knowledge Graph Matching co-located with the 18th International Semantic Web
  Conference (ISWC 2019)},
\byear{2019}.
\url{http://ceur-ws.org/Vol-2553/paper5.pdf}.
\end{bchapter}
\endbibitem

\bibitem{NguyenNIT19}
\begin{barticle}
\bauthor{\binits{P.}~\bsnm{Nguyen}},
\bauthor{\binits{K.}~\bsnm{Nguyen}},
\bauthor{\binits{R.}~\bsnm{Ichise}} and
\bauthor{\binits{H.}~\bsnm{Takeda}},
\batitle{EmbNum+: Effective, Efficient, and Robust Semantic Labeling for
  Numerical Values},
\bjtitle{New Gener. Comput.}
\bvolume{37}(\bissue{4})
(\byear{2019}),
\bfpage{393}--\blpage{427},
\bcomment{DOI: \url{https://doi.org/10.1007/s00354-019-00076-w}}.
\url{https://doi.org/10.1007/s00354-019-00076-w}.
\end{barticle}
\endbibitem

\bibitem{LuoLCZ18}
\begin{bchapter}
\bauthor{\binits{X.}~\bsnm{Luo}},
\bauthor{\binits{K.}~\bsnm{Luo}},
\bauthor{\binits{X.}~\bsnm{Chen}} and
\bauthor{\binits{K.Q.}~\bsnm{Zhu}},
\bctitle{{Cross-Lingual Entity Linking for Web Tables}},
in: \bbtitle{Proceedings of the Thirty-Second Conference on Artificial
  Intelligence (AAAI)},
\bpublisher{{AAAI} Press},
\byear{2018},
pp.~\bfpage{362}--\blpage{369}.
\end{bchapter}
\endbibitem

\bibitem{knoblock2012semi}
\begin{bchapter}
\bauthor{\binits{C.A.}~\bsnm{Knoblock}},
\bauthor{\binits{P.A.}~\bsnm{Szekely}},
\bauthor{\binits{J.L.}~\bsnm{Ambite}},
\bauthor{\binits{A.}~\bsnm{Goel}},
\bauthor{\binits{S.}~\bsnm{Gupta}},
\bauthor{\binits{K.}~\bsnm{Lerman}},
\bauthor{\binits{M.}~\bsnm{Muslea}},
\bauthor{\binits{M.}~\bsnm{Taheriyan}} and
\bauthor{\binits{P.}~\bsnm{Mallick}},
\bctitle{{Semi-automatically Mapping Structured Sources into the Semantic
  Web}},
in: \bbtitle{The Semantic Web: Research and Applications - 9th Extended
  Semantic Web Conference, {ESWC} 2012, Heraklion, Crete, Greece, May 27-31,
  2012. Proceedings},
\bsertitle{Lecture Notes in Computer Science},
Vol.~\bseriesno{7295},
\bpublisher{Springer},
\byear{2012},
pp.~\bfpage{375}--\blpage{390},
\bcomment{DOI: \url{https://doi.org/10.1007/978-3-642-30284-8\_32}}.
\end{bchapter}
\endbibitem

\bibitem{knap2017towards}
\begin{bchapter}
\bauthor{\binits{T.}~\bsnm{Knap}},
\bctitle{{Towards Odalic, a Semantic Table Interpretation Tool in the ADEQUATe
  Project}},
in: \bbtitle{Proceedings of the 5th International Workshop on Linked Data for
  Information Extraction co-located with the 16th International Semantic Web
  Conference {(ISWC} 2017), Vienna, Austria, October 22, 2017},
\bsertitle{{CEUR} Workshop Proceedings},
Vol.~\bseriesno{1946},
\bpublisher{CEUR-WS.org},
\byear{2017},
pp.~\bfpage{26}--\blpage{37}.
\url{http://ceur-ws.org/Vol-1946/paper-04.pdf}.
\end{bchapter}
\endbibitem

\bibitem{cutrona2019asia}
\begin{botherref}
\oauthor{\binits{V.}~\bsnm{Cutrona}},
\oauthor{\binits{M.}~\bsnm{Ciavotta}},
\oauthor{\binits{F.D.}~\bsnm{Paoli}} and
\oauthor{\binits{M.}~\bsnm{Palmonari}},
{ASIA: a Tool for Assisted Semantic Interpretation and Annotation of Tabular
  Data}
\textbf{2456}
(2019),
209--212.
\url{http://ceur-ws.org/Vol-2456/paper54.pdf}.
\end{botherref}
\endbibitem

\bibitem{taheriyan2016learning}
\begin{barticle}
\bauthor{\binits{M.}~\bsnm{Taheriyan}},
\bauthor{\binits{C.A.}~\bsnm{Knoblock}},
\bauthor{\binits{P.A.}~\bsnm{Szekely}} and
\bauthor{\binits{J.L.}~\bsnm{Ambite}},
\batitle{{Learning the Semantics of Structured Data Sources}},
\bjtitle{J. Web Semant.}
\bvolume{37-38}
(\byear{2016}),
\bfpage{152}--\blpage{169},
\bcomment{DOI: \url{https://doi.org/10.1016/j.websem.2015.12.003}}.
\end{barticle}
\endbibitem

\end{thebibliography}

\clearpage

\begin{appendix}

\section{Running Example: RML Definitions}

\begin{minipage}{\textwidth}

\begin{lstlisting}[captionpos=b, caption={A working example of an RML file transforming the data table given in Fig. \ref{fig:example_table} (``sky\_sensors.tsv'') into the data graph indicated in Fig. \ref{fig:correct_mapping_example}. To perform the transformation, the table needs to be pre-processed: the column titles ``col0'',\ldots,``col3'' were added, and a ``rowNumber'' column. For brevity, we skip the mapping of the second and third column to \voc{time}{Interval}.}, label=lst:rml, frame=single, escapeinside={(*}{*)}]
@prefix rr: <http://www.w3.org/ns/r2rml#>.
@prefix rml: <http://semweb.mmlab.be/ns/rm(**)l#>.
@prefix ex: <http://example.com/resource/>.
@prefix csvw: <http://www.w3.org/ns/csv(**)w#>.

ex:File a rml:source ;
 rml:source ex:FileSource ;
 rml:referenceFormulation <http://semweb.mmlab.be/ns/ql#CSV> .
 
ex:FileSource a csvw:Table;
 csvw:url "sky_sensors.tsv" ;
 csvw:dialect [
  a csvw:Dialect;
  csvw:delimiter "(*$\twoheadrightarrow$*)";
 ] .

ex:Mapping0 a rr:TriplesMap ;
 rml:logicalSource ex:File ;
 rr:subjectMap [
  rr:class sosa:Sensor ;
  rr:template "https://www.w3.org/TR/vocab-ssn/Sensor{col3}" ;
 ];

 rr:predicateObjectMap [
  rr:predicate rdfs:label ;
  rr:objectMap [
   rml:reference "col3";
  ]
 ] ;

 rr:predicateObjectMap [
  rr:predicate sosa:madeObservation ;
  rr:objectMap [
   rr:template "https://www.w3.org/TR/vocab-ssn/Observation{rowNumber}";
  ]
 ] .

ex:Mapping1 a rr:TriplesMap ;
 rml:logicalSource ex:File ;
 rr:subjectMap [
  rr:class sosa:Observation ;
  rr:template "https://www.w3.org/TR/vocab-ssn/Observation{rowNumber}" ;
 ] .
\end{lstlisting}

\end{minipage}

\begin{lstlisting}[float=*,captionpos=b, caption={The Turtle file resulting from running the RML file in Listing \ref{lst:rml} to transform the table given in Fig. \ref{fig:example_table} into a data graph. For brevity, we only consider the first three lines of the data table.}, label=lst:rml2, frame=single, escapeinside={(*}{*)}]
<https://www.w3.org/TR/vocab-ssn/Observation0>
  a <http://www.w3.org/ns/so(**)sa/Observation> .

<https://www.w3.org/TR/vocab-ssn/Observation1>
  a <http://www.w3.org/ns/so(**)sa/Observation> .

<https://www.w3.org/TR/vocab-ssn/Observation2>
  a <http://www.w3.org/ns/so(**)sa/Observation> .

<https://www.w3.org/TR/vocab-ssn/SensorS2>
  a <http://www.w3.org/ns/so(**)sa/Sensor>;
  <http://www.w3.org/2000/01/rdf-schema#label> "S2";
  <http://www.w3.org/ns/so(**)sa/madeObservation>
    <https://www.w3.org/TR/vocab-ssn/Observation0> .

<https://www.w3.org/TR/vocab-ssn/SensorS3>
  a <http://www.w3.org/ns/so(**)sa/Sensor>;
  <http://www.w3.org/2000/01/rdf-schema#label> "S3";
  <http://www.w3.org/ns/so(**)sa/madeObservation>
      <https://www.w3.org/TR/vocab-ssn/Observation1>,
      <https://www.w3.org/TR/vocab-ssn/Observation2> .
\end{lstlisting}

\end{appendix}

%

\end{document}